%% file: main.tex
\documentclass[lettersize,journal]{IEEEtran}

\AtBeginDocument{%
  }

\usepackage{amsmath, amssymb, amsfonts}
\usepackage{graphicx}
\usepackage{tabularx, booktabs, makecell, tabularray}
\usepackage{xspace}
\usepackage{xcolor}
\usepackage{enumitem}
\usepackage[caption=false, font=normalsize]{subfig}
\usepackage{algorithm}
\usepackage[noend]{algpseudocode}
\usepackage{lipsum}  % Lorem ipsum text
\usepackage{CJKutf8}
\usepackage{url}
\usepackage{hyperref}

\newcommand{\ie}{\emph{i.e.} }

\newcommand{\eg}{\emph{e.g.} }

\newcommand{\etal}{\emph{et al.} }

\newcommand\figref[1]{Fig.~\ref{#1}}

\newcommand\tabref[1]{Tab.~\ref{#1}}

\newcommand\secref[1]{Sec.~\ref{#1}}

\newcommand{\sysname}{{\sf AdaFlowLite}\xspace}
\newcommand{\sysnameposs}{{\sf AdaFlowLite's}\xspace}

\ifodd 1

\newcommand\rev[1]{\textcolor{black}{#1}}

\newcommand\wfm[1]{\textcolor{black}{#1}}
\newcommand\wyd[1]{\textcolor{black}{#1}}

\else
\fi

\begin{document}
%\definecolor{mypurple}{RGB}{128,0,128}

\title{AdaFlowLite: Scalable and Non-blocking Inference on Asynchronous Mobile Data}

\author{
Sicong Liu,~\IEEEmembership{Member,~IEEE,}
Fengmin Wu, 
Yuan Gao,
Bin Guo*,~\IEEEmembership{Senior Member,~IEEE,}
Zimu Zhou, ~\IEEEmembership{Member,~IEEE,}
Hongkai Wen,~\IEEEmembership{Member,~IEEE,}
Zhiwen Yu,~\IEEEmembership{Senior Member,~IEEE}
\thanks{{*}Corresponding Author: Bin Guo (e-mail:guob@nwpu.edu.cn)}
}

\markboth{Journal of \LaTeX\ Class Files,~Vol.~14, No.~8, August~2021}%
{Shell \MakeLowercase{\textit{et al.}}: A Sample Article Using IEEEtran.cls for IEEE Journals}

\IEEEpubid{0000--0000/00\$00.00~\copyright~2021 IEEE}

\maketitle

\begin{abstract}
The rise of mobile devices equipped with numerous sensors, such as LiDAR and cameras, has driven the adoption of multi-modal deep intelligence for distributed sensing tasks, such as smart cabins and driving assistance. 
However, the arrival time of mobile sensory data vary due to modality size and network dynamics, which can lead to delays (if waiting for slow data) or accuracy decline (if inference proceeds without waiting). 
Moreover, the diversity and dynamic nature of mobile systems exacerbate this challenge.
In response, we present a shift to \textit{opportunistic} inference for asynchronous distributed multi-modal data, enabling inference as soon as partial data arrives. 
While existing methods focus on optimizing modality consistency and complementarity, known as modal affinity, they lack a \textit{computational} approach to control this affinity in open-world mobile environments. 
\sysname pioneers the formulation of structured cross-modality affinity in mobile contexts using a hierarchical analysis-based normalized matrix. 
This approach accommodates the diversity and dynamics of modalities, generalizing across different types and numbers of inputs. 
Employing an multi-modal lightweight Swin Transformer (MMLST), \sysname facilitates real-time and flexible data imputation, adapting to various modalities and downstream tasks without retraining. 
Experiments show that \sysname significantly reduces inference latency by up to 80.4\% and enhances accuracy by up to 62.1\%, while achieving nearly a 50\% reduction in energy consumption, outperforming status quo approaches.
Also, this method can enhance LLM performance to preprocess asynchronous data.

\end{abstract}


\maketitle

\input{body/introduction.tex}

\input{body/motivation}
\input{body/affinity_1}

\input{body/inference}
\input{body/experiments}
\input{body/related}
\input{body/conclusion}

\section*{Acknowledgment}
This work was partially supported by the National Science Fund for Distinguished Young Scholars (62025205) and the National Natural Science Foundation of China (No. 62032020, 62102317,62472354).

% \begin{acks}
% \end{acks}

\bibliography{hmt-base}

\begin{IEEEbiography}[{\includegraphics[width=1in,height=1.25in,clip,keepaspectratio]{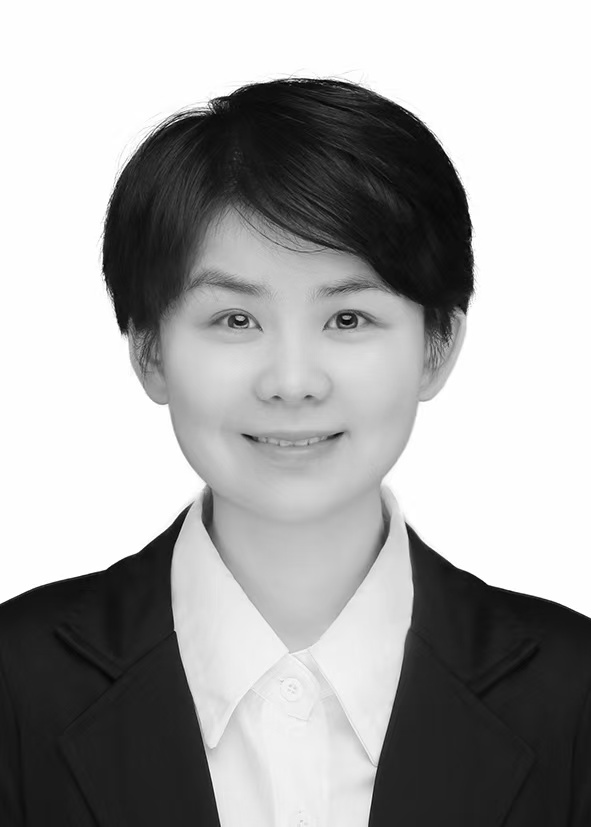}}]{Sicong Liu}
received the PhD degree in school of computer science and technology from Xidian University in 2020. Now she is an associate professor with school of computer science, Northwestern Polytechnical University. Her research interests include mobile computing and resource-efficient mobile deep learning. She has served as the TPC member of MobiSys 2021 and BigCom 2021.
\end{IEEEbiography}

\begin{IEEEbiography}[{\includegraphics[width=1in,height=1.25in,clip,keepaspectratio]{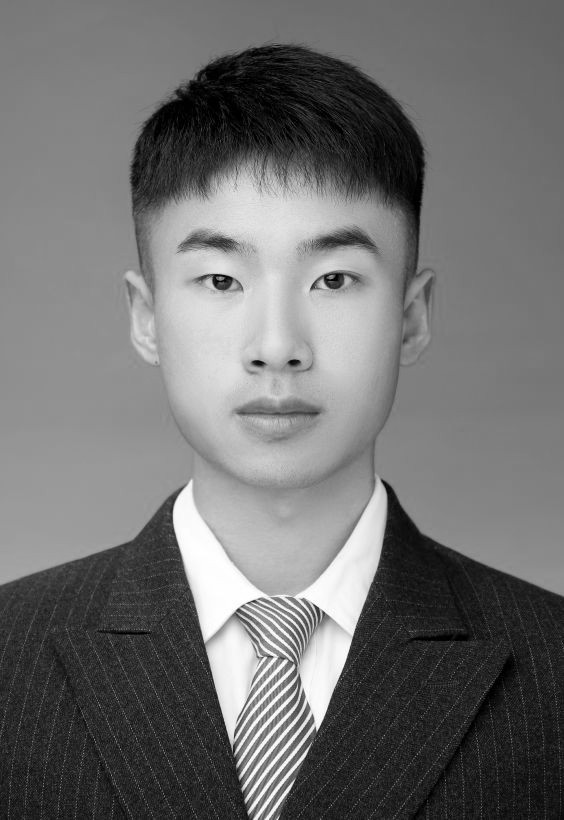}}]{Fengmin Wu}
is a Doctoral student at Northwestern Polytechnical University, China. He received B.E. from China University of Petroleum (Beijing) in 2023. His research interests include resource-efficient mobile deep learning and mobile systems.
\end{IEEEbiography}

\begin{IEEEbiography}[{\includegraphics[width=1in,height=1.25in,clip,keepaspectratio]{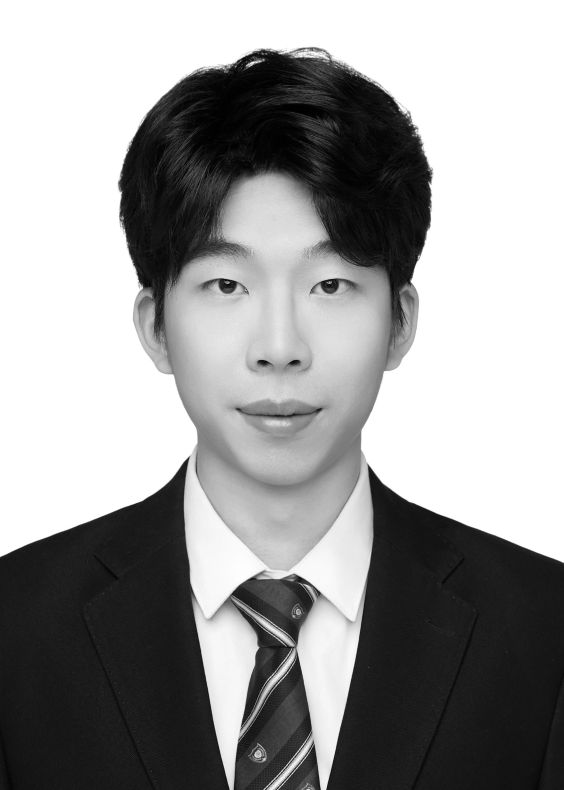}}]{Yuan Gao}
is a Master's student at Northwestern Polytechnical University, China. He received his B.E. from Northwestern Polytechnical University in 2023. His interest is in Heterogeneous Multi-Agent Collaborative Sensing and Perception.
\end{IEEEbiography}

\begin{IEEEbiography}[{\includegraphics[width=1in,height=1.25in,clip,keepaspectratio]{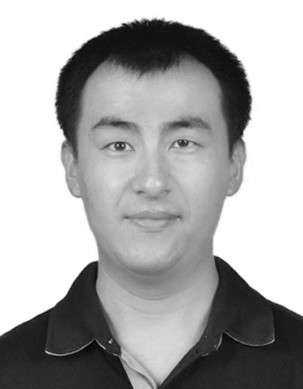}}]{Bin Guo}
received the PhD degree in computer science from Keio University, Japan, and then was a postdoc researcher with Institut Telecom SudParis, France. He is a professor with Northwestern Polytechnical University, China. His research interests include ubiquitous computing, mobile crowd sensing, and HCI. He has served as an associate editor of the IEEE Communications Magazine and the IEEE Transactions on Human-Machine-Systems, the guest editor of the ACM Transactions on Intelligent Systems.
\end{IEEEbiography}

\begin{IEEEbiography}[{\includegraphics[width=1in,height=1.25in,clip,keepaspectratio]{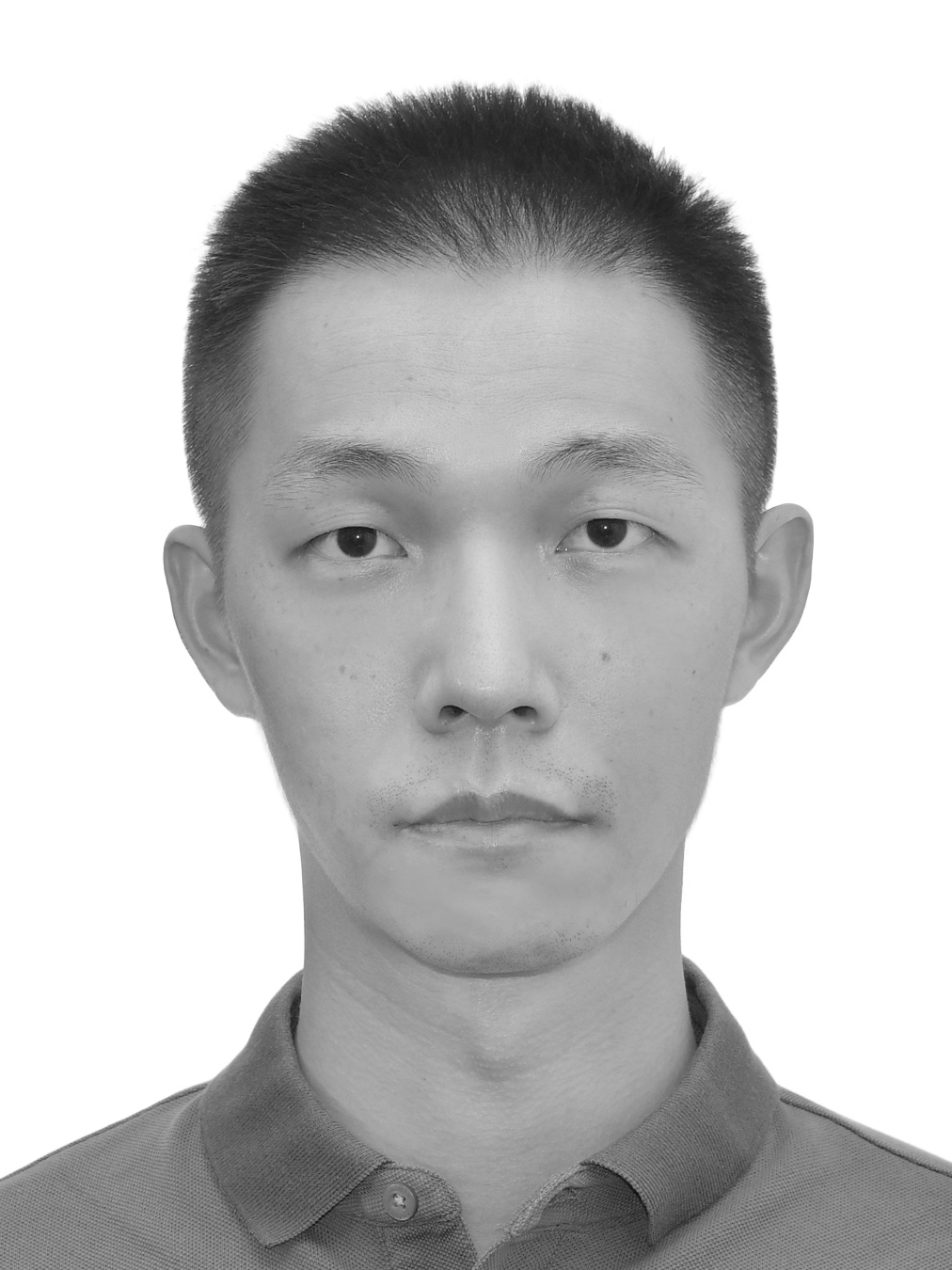}}]{Zimu Zhou}
is currently an assistant professor in the Department of Data Science, City University of Hong Kong. He obtained his Ph.D. from the Hong Kong University of Science and Technology in 2016 and B.E. from Tsinghua University in 2011. His research interest lies broadly in mobile and ubiquitous computing, with a focus on AIoT. zimuzhou@cityu.edu.hk
Zimu Zhou’s research is supported by CityU APRC grant (No. 9610633).
\end{IEEEbiography}

\begin{IEEEbiography}[{\includegraphics[width=1in,height=1.25in,clip,keepaspectratio]{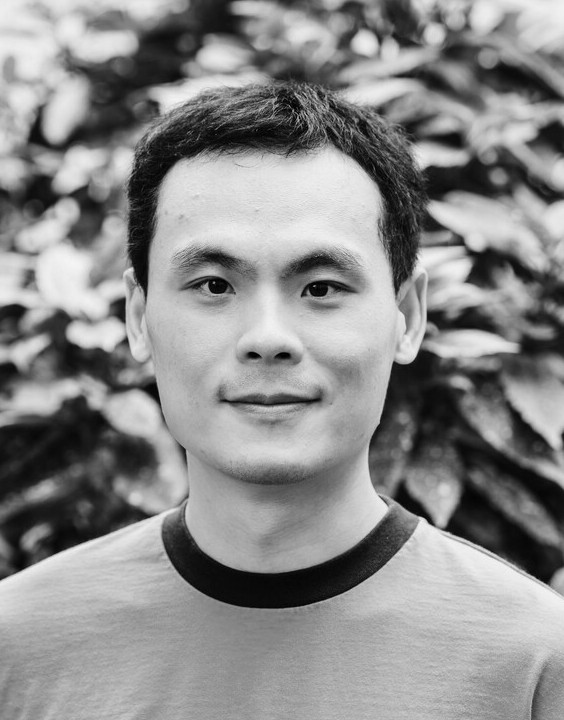}}]{Hongkai Wen}
is a professor (Chair in Machine Learning Systems) at the Department of Computer Science, the University of Warwick, where he lead the AI Systems Lab. He is a Fellow of the Alan Turing Institute, the UK’s national institute for data science and AI. At the Turing, he is also an Independent Scientific Advisor for the BridgeAI programme, and a member of the Turing Research Ethics (TREx) team.
\end{IEEEbiography}

% \vspace{-170mm}
\begin{IEEEbiography}[{\includegraphics[width=0.98in,height=1.25in,clip,keepaspectratio]{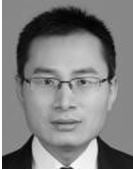}}]{Zhiwen Yu}
received the Ph.D. degree in computer science from Northwestern Polytechnical University, Xi’an, China, in 2005. He is currently the vice president of Harbin Engineering University and a professor at the School of Computer Science, Northwestern Polytechnical University. He was an Alexander Von Humboldt Fellow with Mannheim University, Germany, and a Research Fellow with Kyoto University, Kyoto, Japan. His research interests include ubiquitous computing, HCI, and mobile sensing and computing.
\end{IEEEbiography}

\end{document}

%% file: body/introduction.tex
\section{Introduction}
\label{sec:intro}
The broad integration of low-power, rich-data sensors like cameras, LiDAR, acoustic sensors, hyperspectral cameras, and radio frequency detectors has greatly advanced the sensing potential of mobile devices~\cite{bib:mml:b,bib:er:l}.
Leveraging multiple sensors for multi-modal inference provides robust perception outcomes compared to single-sensor systems across various domains, such as smart cockpits (\eg Google automotive service~\cite{bib:gas:g}), 
driving assistance (\eg Tesla Autopilot~\cite{bib:tesla:t}),
3D scene understanding (\eg Microsoft Azure Kinect DK~\cite{bib:kinect:m}), health monitoring (\eg Apple HealthKit~\cite{bib:apple:a}), and AR/VR (\eg Meta Oculus Rift~\cite{bib:face:a}).
The advantages of \textit{distributed} multi-modal systems are particularly evident 
in challenging environments or objects. 
For instance, observing occluded objects, penetrating through walls, operating in low-light conditions, and navigating around corners. 
The diversity of sensor data from various fields of view and sensitivities supports independent cross-validation from multiple vantage points and enables detailed extraction of complex patterns (\eg location, direction, and materials).

\wfm{In mobile and distributed computing systems, a fundamental challenge lies in efficiently processing and analyzing massive real-time multi-modal data streams generated by advanced sensors, especially in vehicle-/drone-mounted scenarios (we defer more detail in \secref{gan}).
Many studies focus on technologies and tools for handling multi-modal data, such as adopting AdaptSegNet for fusing complex signals~\cite{bib:ssm:v}, or further explored partitioning models across mobile devices and the cloud to enhance processing speed~\cite{bib:neu:k}.
However, the significant \textit{gap} exists in open-world mobile deployments, characterized by input \textit{asynchrony} and \textit{heterogeneity}:}

%工业界已经在解决这个问题了，
% \begin{itemize}
\wfm{\noindent$\bullet$ \textbf{Asynchronism}. In distributed multi-modal systems, particularly those employing vehicle-/drone-mounted sensors, the diversity of input modalities (\eg video $\textit{vs.}$ LiDAR) coupled with dynamic network conditions often leads to asynchronous data arrivals at the fusion device. 
These challenges are further compounded by the resource constraints typical of mobile systems, such as limited computational power and the absence of clock synchronization among low-power mobile sensors. 
These factors can result in delays when the system waits for slower data streams or a decline in accuracy if fusion proceeds without these data streams being synchronized.
Consider a real-world example involving an autonomous driving system equipped with six cameras and one LiDAR sensor. 
In this scenario, an asynchronous delay of approximately 40ms may occur for LiDAR data over a 100Mbps bandwidth, given a data size ratio of $1:4$ between LiDAR and camera data. This delay can increase the total latency from 1s to 4.3s for processing 81 frames within a $4s$ window when the system chooses to wait for the slower LiDAR data. Alternatively, proceeding without waiting for data synchronization can degrade the fusion accuracy by 49.7\%. 
To maintain operational safety and efficiency, mobile systems must minimize latency, such as $\leq 26$ms per frame~\cite{bib:dl:m}.
We defer more details to \secref{problem1}.}

\wfm{\noindent$\bullet$ \textbf{Heterogeneity}. In distributed multi-modal systems, particularly those with vehicle-/drone-mounted sensors, variations in the field of view, sensitivity, noise (e.g., lighting conditions), and feature distribution significantly influence the \textit{consistency} and \textit{complementarity} of data fusion outcomes~\cite{bib:deep:l}.
\IEEEpubidadjcol
The effective management of these factors is crucial when deciding whether to discard or impute slow-arriving modalities, as these decisions directly impact inference accuracy and system latency.
For instance, we find that discarding 50\% of the slow data, which exhibits approximately 33\% arrival delays, results in a 28.1\% decrease in accuracy using the real-world autonomous driving dataset, nuScenes~\cite{bib:nuscenses:ca}. 
Employing the K-Nearest Neighbors (KNN) approach~\cite{bib:knn:c}, a common method for asynchronous data imputation for addressing this problem still leads to a 9.7\% reduction in accuracy and a 328\% increase in latency. 
Additional details on this observation are presented in \secref{subsec:observation}.}

\wfm{As a separate note, cross-modal \textit{consistency} and \textit{complementarity}, key components of what we describe modality \textit{\textbf{affinity}}, are crucial for enhancing inference robustness and accuracy by leveraging \textit{modality-shared} and \textit{modality-specific} information, respectively~\cite{bib:cosmo:oyxm}.
In diverse mobile scenarios, the addition of sensors necessitates a \textit{refined cross-modality affinity analysis}.
Current affinity analysis often relies on \textit{complex manual} adjustments and significant computational overhead, as detailed in~\secref{problem2}.
Moreover, modality affinity is subject to fluctuations due to input \textit{asynchronism} and \textit{heterogeneity}, which vary not only across tasks but also with identical inputs, as observed in~\secref{subsec:observation}.
Considering these characteristics, both existing synchronous (blocking) methods~\cite{bib:mml:b, bib:multimodal:hong, bib:multimodal:liu} and asynchronous (non-blocking) approaches~\cite{bib:lls:L,bib:echo:l,bib:adaflow:wu} encounter significant challenges as below:}

\wfm{\noindent$\bullet$ \textbf{\textit{Challenge $\#$1}}: 
Most multi-modal methods balance accuracy and latency by enhancing data \textit{consistency} and \textit{complementarity}.
However, these methods often only provide qualitative affinity analysis under ideal conditions~\cite{bib:taskonomy:AZ, bib:prov:z}.
And both qualitative or quantitative method(\ie the preliminary AdaFlow~\cite{bib:adaflow:wu}), are primarily designed to handle the \textit{reduction} (\ie missing subsets) of modalities but fail to accommodate the \textit{addition} of sensor types or numbers. 
Our approach utilizes a \textit{prior-free modality affinity refinement} that quantitatively analyzes cross-modal affinity and scaly accommodate to new modalities. This facilitates the effective management of diverse and dynamic inputs in systems equipped with vehicle-/drone-mounted sensors.}

\wfm{\noindent$\bullet$ \textbf{\textit{Challenge $\#$2}}: 
Scheduling asynchronous and heterogeneous multi-modal data for low-latency inference on resource-constrained mobile platforms such as vehicle-/drone-mounted systems is non-trivial. 
It involves NP-hard multi-choice, multi-strategy optimization problems~\cite{bib:taskonomy:AZ, bib:adam:k, bib:rmsprop:t}. 
Resource limitations inherent to mobile environments make the task even more challenging.
Moreover, designing a lightweight one-fit-all data imputation module that can handle \textit{dynamic} and \textit{diverse} input modalities, various subsequent tasks, and fluctuating asynchronous delays is essential yet challenging. }

\wfm{To address the challenges of asynchronous data integration in mobile environments with resource constraints, we propose a shift to \textit{opportunistic} inference.
The server (\eg on a vehicle-/drone-mounted mobile device) initiates inference processes as soon as any partial data becomes available, optimizing the use of limited computational resources. 
This method is particularly effective in scenarios where data collection is subject to variable delays and interruptions, common in mobile settings.
\sysname leverages fully computational \textit{modality affinity} to ensure robust data fusion at runtime, that not only models but also efficiently optimizes inference activities based on the affinity between different data modalities. }
% This approach allows for efficient processing even under the stringent power and processing limitations typical of mobile platforms deployed in field operations.

\wfm{\noindent$\bullet$ \textit{\textbf{First}}, inspired by~\cite{bib:taskonomy:AZ}, which investigates \textit{task} affinity using \textit{unimodal} (visual) data during \textit{learning}, we extend this to the \textit{modality} level during \textit{inference}, enabling dynamic adaptation across diverse modalities.
\sysname leverages t-SNE combined with the Analytic Hierarchy Process (AHP) to normalize the modality affinity matrix.
To not only handle \textit{missing} modalities but also incorporate new \textit{unseen} sensors/modalities, \sysname integrates a \textit{prior-free affinity refinement} module, enabling seamless integration into the structured modality affinity matrix. 
This ensures robust decision-making in dynamic data availability and diverse mobile scenarios.}

\wfm{\noindent$\bullet$ \textit{\textbf{Second}}, to enable low-latency non-blocking inference on resource-limited devices that adapts to varying input modalities and types \textit{without retraining} at runtime, we introduce a \textit{one-fit-all multi-modal lightweight Swin Transformer} (MMLST). 
% This model is specifically designed for mobile computing scenarios where computational resources are limited. 
It adaptively processes diverse input features and modalities and finely controls modal affinity in dynamic opportunistic inference. 
Notably, the proposed MMLST is \textit{lightweight} and suitable for resource-constrained mobile devices such as vehicle-/drone-mounted systems.}

We evaluate the performance of \sysname on real-world 3D object recognition tasks and scenarios with diverse data heterogeneity and asynchrony. 
Results show a reduction of up to 80.4\% in inference time with an improvement of up to 62.1\% in accuracy, while achieving nearly a 50\% reduction in energy consumption.
Especially, it displays the affinity matrix enables opportunistic inference based on different downstream tasks and resource-constrained scenarios.
This method can also be a pre-module for LLM, enhancing performance on asynchronous data.
Our main contributions are as follows.
\begin{itemize}
    \item As far as we know, this is the first work to integrate quantitative modality affinity control into distributed multi-modal inference, addressing \textit{system asynchrony} and \textit{diverse, dynamic demands}.
    It eliminates waiting times for slow data and minimizes accuracy loss when excluding such data.
    \item We introduce \sysname, a system for opportunistic inference that adapts to asynchronous and heterogeneous data streams. 
    It harnesses structured and scalable affinity control to enable adaptive fusion of dynamic data. 
    Also, \sysname proposes a lightweight MMLST to achieve precise and timely data imputation on mobile devices.
    
    \item Experiments show that \sysname outperforms existing a-/synchronous methods~\cite{bib:patch:w,bib:mdl:n,bib:pcn:Y,bib:knn:c,bib:adaflow:wu} in trading off between inference accuracy, latency and energy consumption across various real-world mobile tasks, inputs, and scenarios. 
\end{itemize}

%% file: body/motivation.tex
\section{Overview}

\subsection{Problem Analysis}
\label{subsec:problem}

% The server-based fusion scheme integrates multiple mobile devices, each equipped with one or more sensors, to fuse distributed sensor data at a central server for inference. 
% This scheme is common in mobile applications. 
% For example, smartphones, watches, and earphones transmit data from various sensors, including accelerometers and GPS, to facilitate health and activity monitoring~\cite{bib:avec:v}. 
% Similarly, vehicle cabins utilize internal and external radar and cameras to observe cabin conditions and external road safety~\cite{bib:vip:shi}.

% \subsection{Challenges}
% \label{subsec:challenge}

\subsubsection{Data Asynchronism Problems}
\label{problem1}
\begin{figure}[t]
    \centering 
    % \vspace{-3mm}
    \subfloat[]{\label{fig:exp_alpha}
    \includegraphics[height=0.26\linewidth]{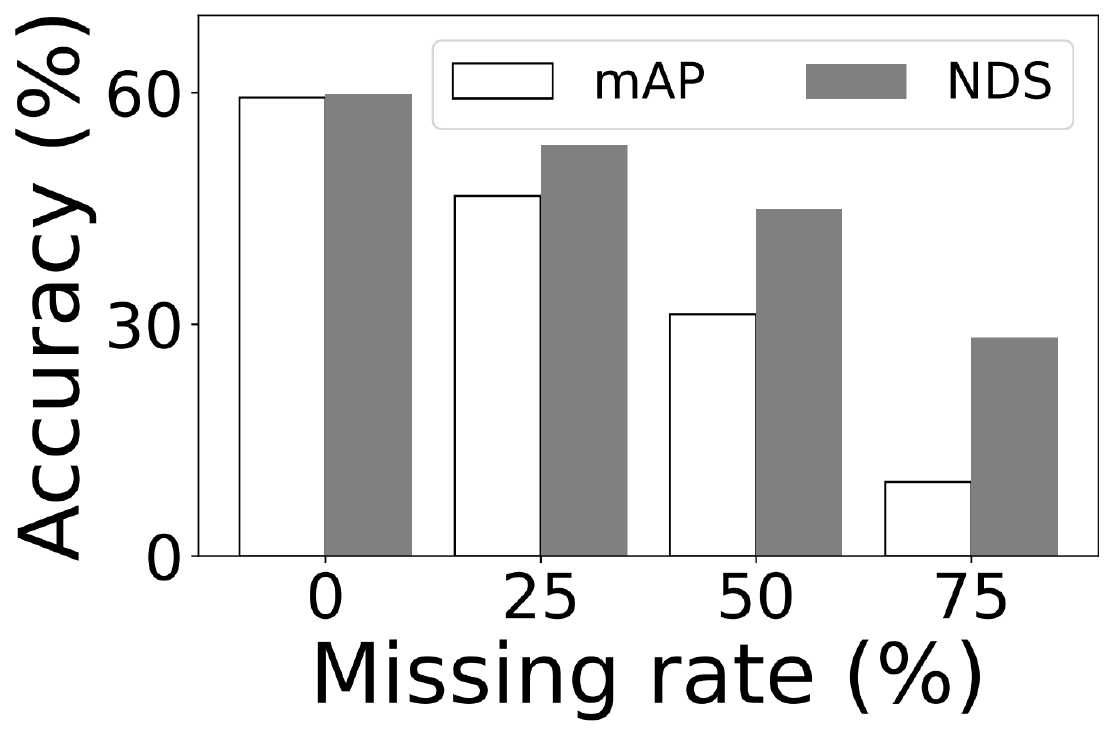}}
    \hspace{3mm}
     \subfloat[]{\label{fig:exp_lr}
\includegraphics[height=0.26\linewidth]{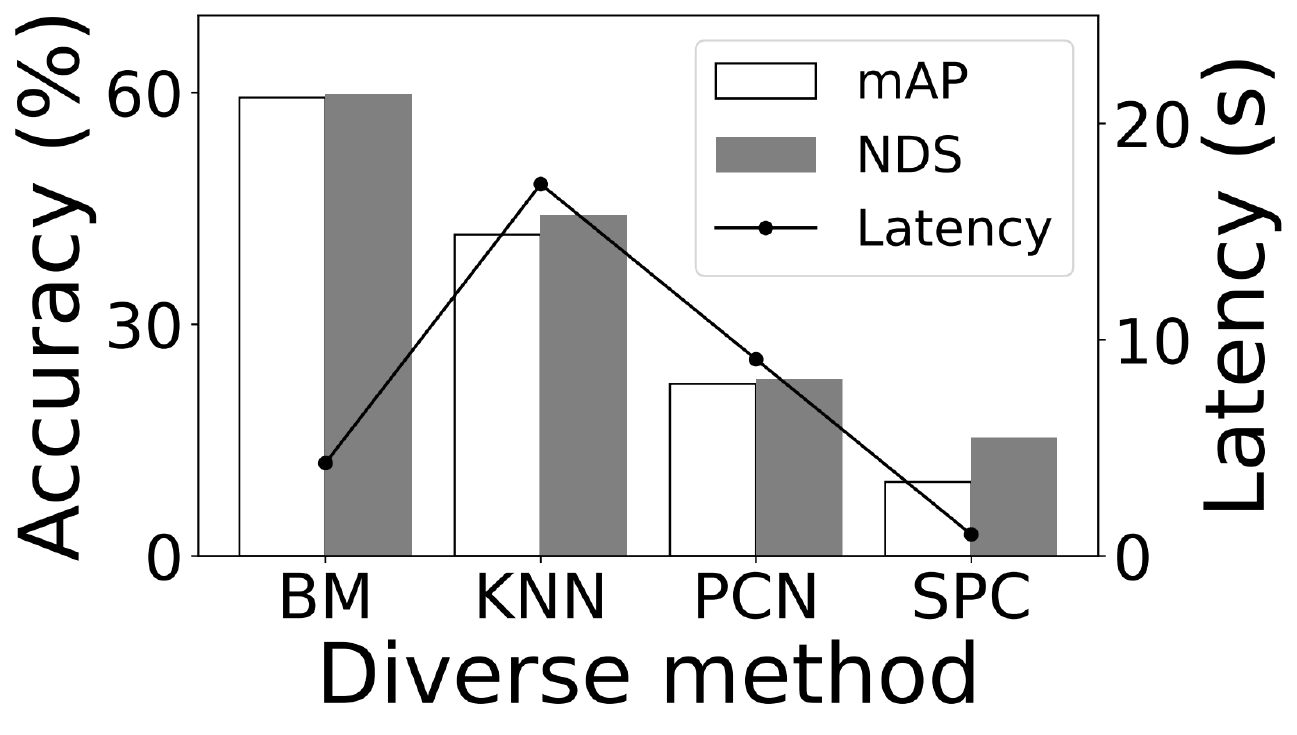}}
    %\vspace{-3mm}
    \caption{(a) Inference accuracy, (b) accuracy \& latency of existing syn-/asynchronous methods, under varying missing rates of slow modality (\ie 3D point cloud).}
    % \vspace{-2mm}
\label{fig-motivation}
\end{figure}

To analyze and observe the motivation, we conducted the following experimental tests.
We test a distributed multi-modal 3D object recognition system on nuScenses dataset~\cite{bib:nuscenses:ca}.
nuScenses, collected from real-world autonomous driving scenarios, includes 6 camera and 1 LiDAR views. 
Given the 1:4 data size ratio between LiDAR and camera data, a 40ms asynchronous delay occurs in LiDAR data on a 100Mbps bandwidth, a common scenario in vehicle-mounted and drone-mounted applications where bandwidth is limited and data throughput must be optimized.
% 存在慢速和快速数据的问题。要让人信服。 慢速和快速之间有多少毫秒delay。
\textit{First}, as illustrated in \figref{fig-motivation}a, we simulate data \textit{asynchronism} using diverse missing rates of slow modality (\ie 3D LiDAR point cloud), \ie 0\%, 25\%, 50\%, and 75\%. 
As the missing rate of the slow modality increases, while the fast modality (\ie RGB images) remains fully available, if not waiting for slow data, the inference accuracy progressively declines (\eg 49.8\%).
If waiting for slow data, the latency increases from 1s to 4.3s for 81 frames.
\textit{Second}, as shown in \figref{fig-motivation}b, we test various imputation methods for \textit{asynchronous fusion} such as SPC (Sparse Point Cloud)~\cite{bib:mdl:n}, KNN (K-Nearest Neighbors) \cite{bib:knn:c}, and PCN (Point Completion Net) \cite{bib:pcn:Y}, along with the \textit{synchronous} fusion method BM (Blocking Mechanism)~\cite{bib:patch:w}, across missing rates of 0\%, 25\%, 50\%, and 75\% in the slow data (\ie 3D point cloud) when the fast data (\ie RGB image) is available.
We find that while the blocking method offers the highest accuracy, its latency, nearly equal to the sampling interval of the test data (4s), is too high. Conversely, the sparse point cloud method is the fastest but yields an unacceptable accuracy drop (\ie 4.29\%).
% In summary, none of existing a-/synchronous methods can achieve the best accuracy-latency tradeoff in distributed mobile scenarios with dynamic modality asynchrony and heterogeneity.

\wfm{\subsubsection{Sensor Scalability Problem}
\label{problem2}
Multi-modal systems commonly \textit{assume} that the total number of sensors remains \textit{constant} during operation, a foundational premise in existing frameworks~\cite{bib:mml:b}. However, in diverse mobile scenarios where sensors need to be \textit{added}, whether to refine sensing domains or enhance capabilities, integrating \textit{new} and potentially \textit{unseen} sensors becomes a significant challenge.
This process requires \textit{refining the cross-modality affinity analysis} to accurately establish relationships between new and existing sensors/modalities for effective fusion. 
However, current affinity analysis often relies on \textit{complex manual} derivations and computational overhead~\cite{bib:reid:lu}, making it both time-consuming and impractical for non-experts to implement. 
The challenge intensifies in scenarios such as foldable or flexible grippers~\cite{bib:gripper:u}, where dynamic, real-time modality affinity analysis is essential but remains highly challenging.
Existing methods, whether qualitative~\cite{bib:prov:z} or quantitative (\ie the preliminary AdaFlow~\cite{bib:adaflow:wu}), are primarily designed to handle the \textit{reduction} (\ie missing subsets) of modalities but fail to accommodate the \textit{addition} of sensor types or numbers. 
This limitation exposes a critical gap in enabling \textit{scalable} sensor and modality integration (\ie both scaling up and down), an essential capability for diverse multi-modal systems.}

\subsection{Observations and Opportunities}
\label{subsec:observation}
In mobile applications, especially in resource-constrained environments like \wyd{vehicle-/drone-mounted systems}, real-time distributed multi-modal data fusion is critical for balancing inference accuracy and latency. 
A key challenge of existing methods (discussed in \secref{sec:related}) is their inability to quantitatively assess and manage \textit{modality affinity}, which encompasses both \textit{consistency} and \textit{complementarity} across modalities. 
This gap often results in delays or accuracy drops, exacerbated by the inherent asynchrony and heterogeneity of mobile contexts. 
Despite its importance, modality affinity remains a \textit{black box}.

% In the context of mobile applications, the goal of distributed multi-modal data fusion is to achieve the optimal tradeoff between inference accuracy and latency.
% The key challenge in existing synchronous/asynchronous fusion methods (as shown in \secref{subsec:challenge}) is the lack of quantitative assessment and management of \textit{modality affinity} (\ie consistency and complementarity).
% This leads to fusion delay or accuracy degradation.
% Despite the significance of modality affinity, it remains a black box, further complicated by unpredictable modality asynchrony and heterogeneity under distributed and dynamic mobile conditions. 
% To decipher this complexity, we conduct specific tests and make the following observations.

% \begin{figure}[t]
% 	\centering
% 	\includegraphics[width=0.9\linewidth]{figure/TOP.pdf}
%         \vspace{-3mm}
% 	\caption{Top LiDAR selection.}
%         \vspace{-3mm}
% 	\label{full}
% \end{figure}
% \begin{figure}[t]
% 	\centering
% 	\includegraphics[width=0.9\linewidth]{figure/NON-TOP.pdf}
%         \vspace{-3mm}
% 	\caption{Non-Top LiDAR selection.}
%         \vspace{-3mm}
% 	\label{partial}
% \end{figure}
%扩展
\subsubsection{Modality-specific Modality Affinity}
\begin{figure}[t]
    \centering 
    % \vspace{-3mm}
    \subfloat[]{\label{full}
    \includegraphics[height=0.4\linewidth]{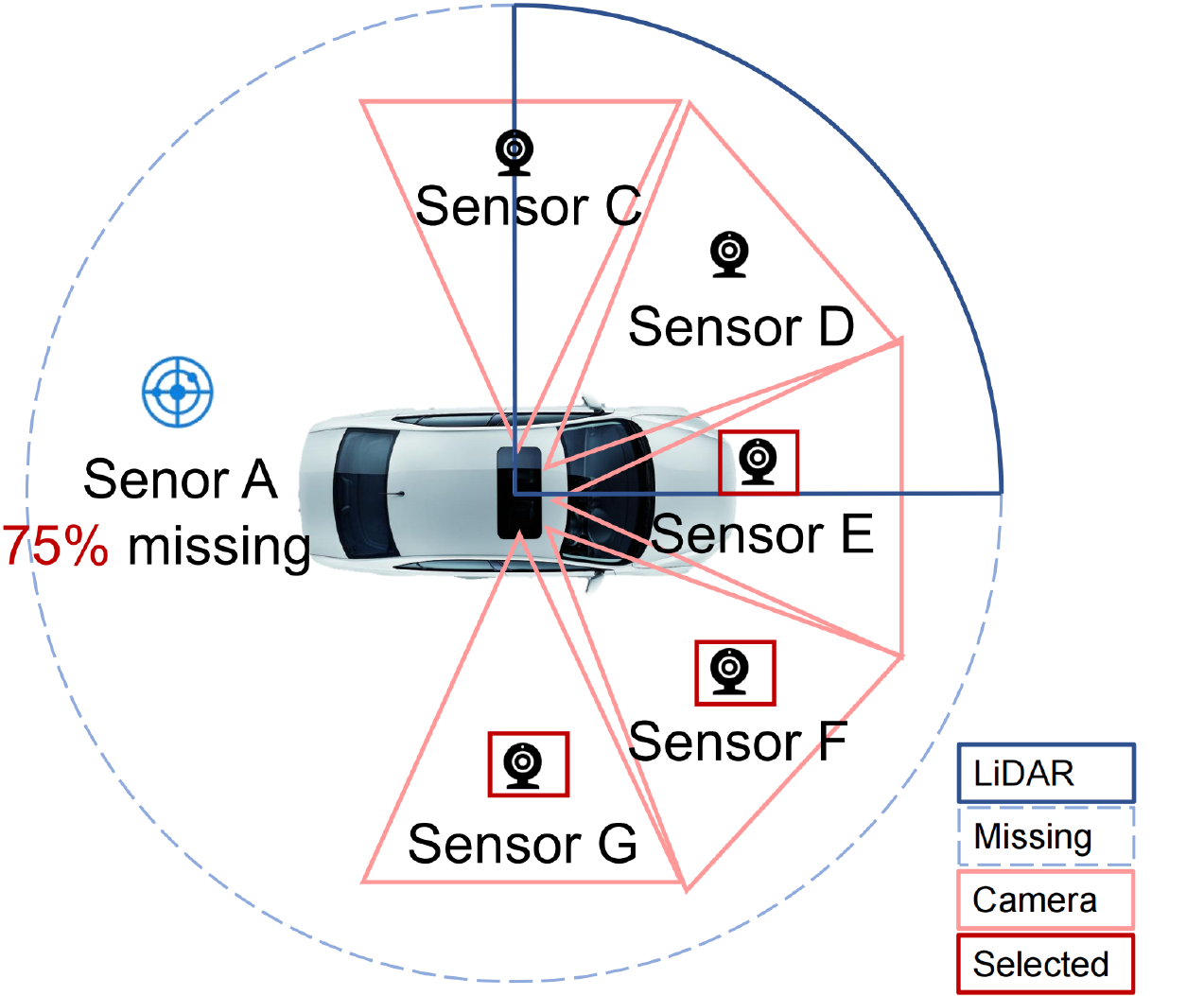}}
     \subfloat[]{\label{partial}
    \includegraphics[height=0.4\linewidth]{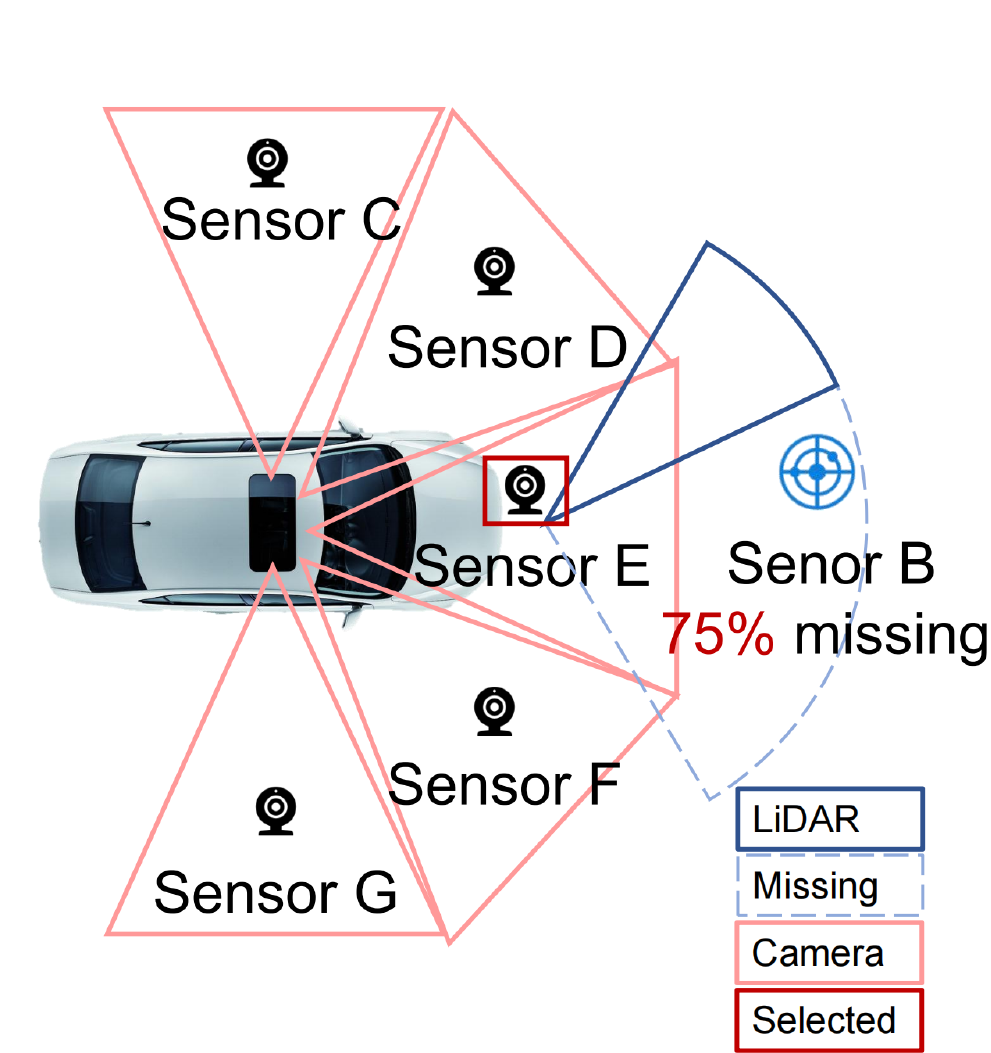}}
    %\vspace{-3mm}
    \caption{(a) Top LiDAR emphasizes complementarity and (b) Non-Top LiDAR emphasizes consistency.}
    % \vspace{-4mm}
\label{fig:lidar}
\end{figure}

Different sensors on mobile devices exhibit distinct modality affinities due to their varying view and size, affecting how they interact with data from other modalities, emphasizing consistency and complementarity. 
For instance, consider a vehicle equipped with seven sensors including top and front LiDARs (Sensor $A$ and $B$), and five positioned cameras (sensor C$\sim$G). 
As illustrated in \figref{full}, when the top full-view LiDAR sensor $A$ (slow data with larger size) misses 75\% of its data, selecting data from sensors C$\sim$G enhances \textit{complementarity} and broadens perspectives, thus improving inference accuracy (\ie from 9.6\% to 54.9\%). 
Conversely, \figref{partial} shows that when the front partial-view LiDAR lags 75\% of its data due to network issues, prioritizing data from sensors C$\sim$G for greater \textit{consistency}, rather than \textit{complementarity}, yields better accuracy. 
This is because the partial-view LiDAR provides specific directional data, necessitating supplementary directional data from other cameras to maintain noise-robust results.
This indicates that the types and number of available fast modalities change dynamically in asynchronous systems. Consequently, modality selection for fusion must adapt dynamically to maintain an optimal performance\wyd{(the necessity analysis discussed in \ref{graph} and the experimental result is in \ref{val}).}

\begin{figure*}[t]
	\centering
	\includegraphics[width=0.75\linewidth]{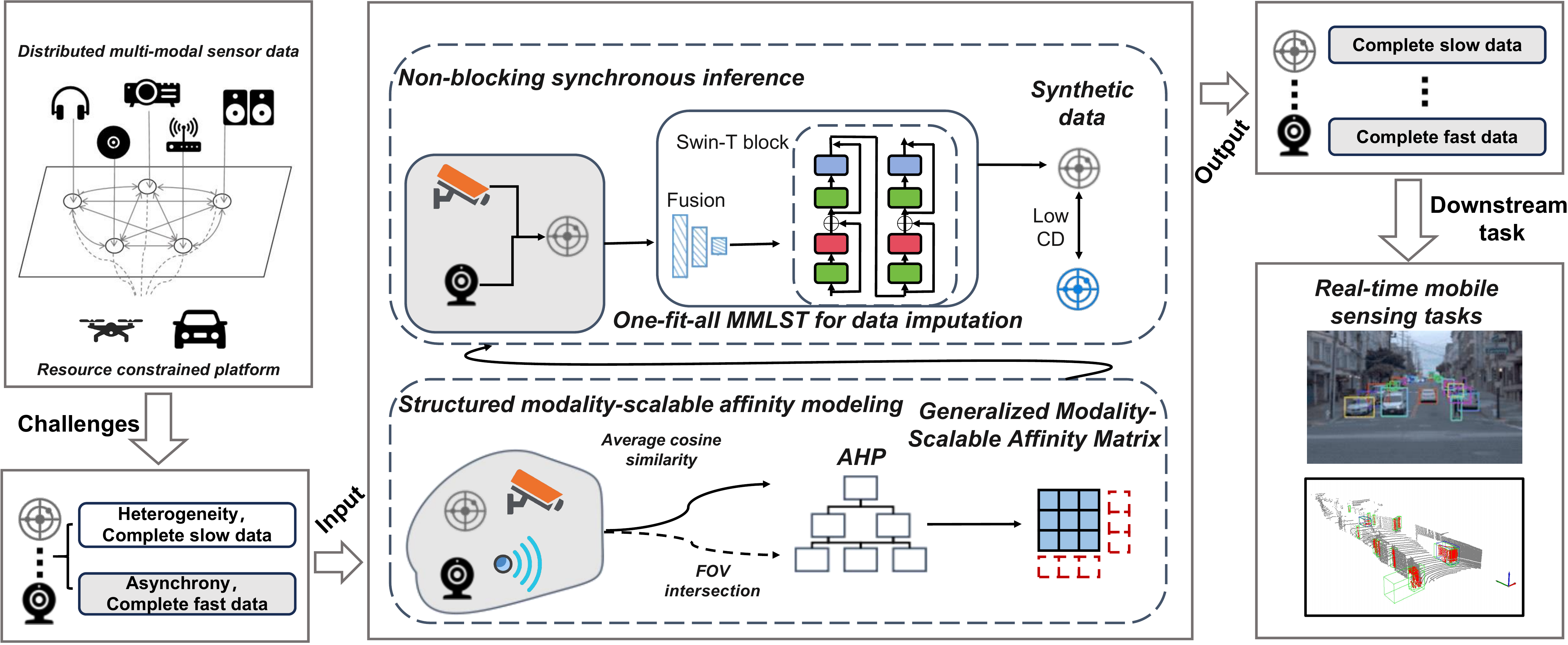}
        % \vspace{-2mm}
	\caption{\wfm{Workflow of \sysname.}}
        % \vspace{-2mm}
	\label{fig:fw}
\end{figure*}

\begin{figure}[t]
	\centering
	\includegraphics[width=0.86\linewidth]{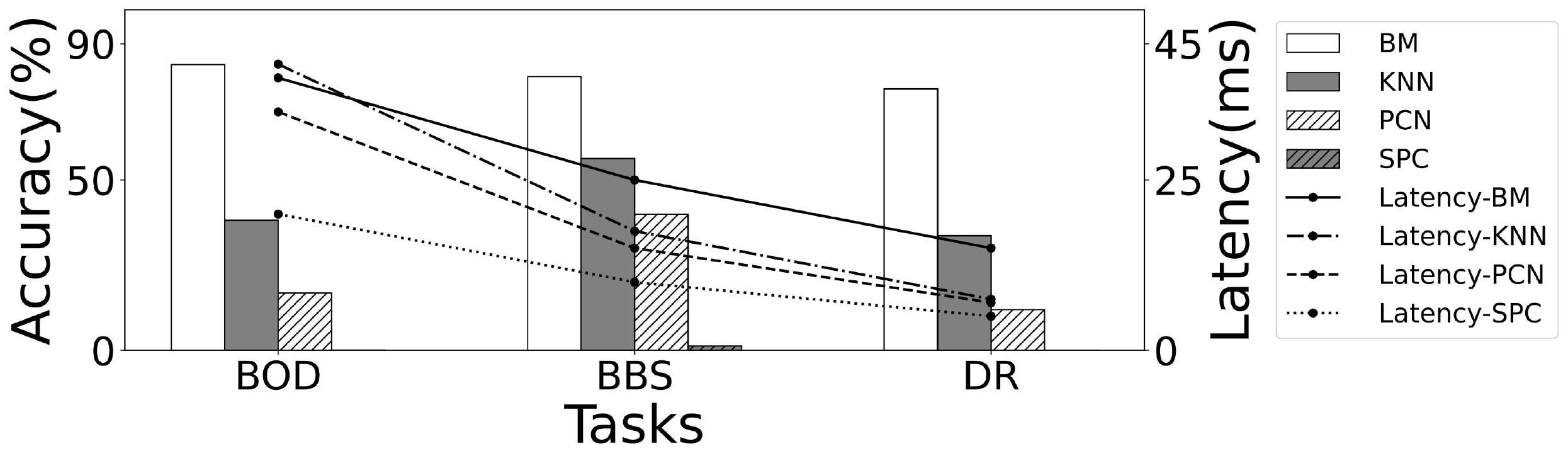}
        \vspace{-2mm}
	\caption{Inference latency and accuracy of BOD, BBS, DR tasks in existing a-/synchronous methods.}
    % \vspace{-2mm}
	\label{fig:vt3}
\end{figure}

\subsubsection{Subsequent Task-specific Modality Affinity}. 
Even the same multi-modal sensor arrival patterns can exhibit varying modality affinities across different \textit{downstream tasks}. 
As demonstrated in \figref{fig:vt3}, we adopt the same heterogeneity and asynchrony settings, \ie 70ms delays of each frame, corresponding to 50\% missing rate, to fairly test three diverse downstream tasks: \textit{i)  BEV object detection (BOD)}, which involves object recognition from a bird's eye view; ii) \textit{Bounding box segmentation (BBS)}, which uses a rectangular box to represent target objects in images or point clouds for semantic segmentation; \textit{iii) Direction recognition (DR)}, which compares direction similarity between the target and the actual orientation. 
%
% The results show that using a naive synchronous method (\ie BM~\cite{bib:patch:w}) leads to significant variations in inference latency across three tasks, \ie 40ms, 25ms, and 15ms for BOD, BSS, DR task, respectively.
% This is because different downstream tasks shape diverse affinity criteria, and even synchronous data will have different inference latency. This makes it difficult to apply these three downstream tasks to latency-sensitive scenarios.
The results show that employing a naive synchronous method (\ie BM~\cite{bib:patch:w}) results in significant variations in latency across three tasks, \eg 40ms for BOD, 25ms for BSS, and 15ms for DR task. 
This variance is due to the distinct affinity criteria shaped by downstream tasks, leading to disparities in inference latency especially when data arrives synchronously. 
Such variability is prohibitive in latency-sensitive applications.
%解释实际的原因，比如什么数据慢多少，对于什么识别带来什么影响
%
Moreover, different data imputation techniques for asynchronous fusion (\ie SPC~\cite{bib:mdl:n}, KNN~\cite{bib:knn:c}, PCN~\cite{bib:pcn:Y}) result in even more pronounced variations across three tasks.
For example, when using PCN asychronous method, the inference latency across three tasks are 35ms, 15ms, 7ms, respectively. 
While accuracy are 16.8\%, 39.9\%, and 11.9\%, respectively. 

These observations demonstrate that \textit{dynamics} in mobile sensing patterns, driven by modality asynchrony, heterogeneity, or subsequent tasks, desired \textit{generalized} modality affinity criteria. 
Building this criteria has notable values.

\wfm{\subsection{Solution Overview}
\label{sec:solution}
To address the above challenges in mobile distributed multi-modal inference systems, we introduce \sysname (\secref{sec:intro}). 
\figref{fig:fw} shows the architecture of \sysname\wyd{(\figref{fig:flow} shows the idea of \sysname)}, which includes two key modules: Structured Modality-Scalable Affinity Modeling and Lightweight Non-blocking Asynchronous Inference.
\textit{First}, the Structured Modality-Scalable Affinity Modeling module (\secref{matrix}), central to \sysname, tackles \textit{Challenge \#1} by constructing a \textit{generalized} modality-scalable affinity matrix. This matrix is specifically designed to manage the inherent \textit{asynchrony} and \textit{heterogeneity} of different modalities, which are common in complex sensing environments. Furthermore, \sysname incorporates a prior-free affinity refinement module for seamlessly integrating new sensors into the structured modality affinity matrix by utilizing \textit{prior-free analysis}, a critical feature in diverse and dynamic mobile settings where prior knowledge of sensor characteristics is often unavailable. 
Also, these capabilities are crucial for dynamically assessing input modalities, particularly in resource-constrained platforms such as vehicles or drones, ensuring robust performance and adaptability under diverse mobile scenarios.
\textit{Second}, the Lightweight Non-blocking Asynchronous Inference module (\secref{sec:inference}) addresses \textit{Challenge \#2} by introducing the multi-modal lightweight Swin Transformer (MMLST). 
This model adapts to various input/output modalities without retraining at runtime, ensuring real-time, accurate inference through dynamic modality fusion. 
It also uses a lightweight dynamic programming approach to select optimal affinity sub-graphs from the matrix, enhancing performance in mobile systems with limited resources.}

%% file: body/affinity_1.tex
\section{Structured Modality-Scalable Affinity Modeling}
\label{matrix}

\subsection{Preliminary of Generalized Cross-modality Affinity Matrix}
\label{paam}

In mobile scenarios characterized by the diversity and dynamics of distributed multi-modal data, which often vary in views and arrival times, ensuring robust decision-making under fluctuating data availability and diverse conditions presents significant challenges. A \textit{generalized modality affinity matrix} is essential to address these challenges by managing the \textit{heterogeneity} and \textit{asynchrony} of data. This matrix enhances \sysnameposs ability to capture complex, nonlinear relationships in high-dimensional data, thereby accommodating diverse tasks alongside dynamically changing input data modalities that are non-stationary, \textit{asynchronous}, and \textit{heterogeneous}.

In a preliminary version of \sysname, namely AdaFlow~\cite{bib:adaflow:wu}, we introduced a mathematical framework called the \textbf{\textit{affinity matrix}}. This framework is designed to tackle the challenges of modality heterogeneity, asynchrony, and varied views within mobile scenarios through a structured two-step strategy.

\wyd{
In the first step, AdaFlow employs the t-SNE method~\cite{bib:t-sne:vd}, which maps Euclidean distances between high-dimensional data points into joint probabilities that reflect their similarities. This process projects heterogeneous modalities into a high-dimensional feature space and excels at information-lossless dimensionality reduction, essential for preserving data integrity. Subsequent to this reduction, AdaFlow calculates pairwise modalities' affinity, ensuring accurate modality fusion. Although obtaining pair-wise modalities' affinity using t-SNE is a straightforward procedure, applying this algorithm globally (across diverse and dynamic mobile setting) and incorporating every single pair-wise modalities' affinity into a unified framework transforms the problem into a NP-hard problem.
}

\wyd{
In the second step, AdaFlow addresses the diversity and dynamics of multi-modal fusion by utilizing the \textit{\textbf{Analytic Hierarchy Process (AHP)}}. AdaFlow are not aiming for a globally optimal solution. 
Instead, it starts with slow data for planning, which reduces the NP-hard problem to a multi-objective programming problem that is more suitable for AHP. The affinity matrix in AdaFlow has three layers: the \textit{Focus}, \textit{Criteria}, and \textit{Alternative} layers. 
The \textit{Focus} layer prioritizes sensors with lower data generation rates or higher vulnerability to lagging, while the \textit{Alternatives} layer includes sensors that transmit data to the terminal first. 
The \textit{Criteria} layer considers two main factors: the average cosine similarity from the t-SNE visualization and the Field Of View (FOV) intersection between different sensors. 
}

\wyd{
To illustrate how an affinity matrix is established, there is a simple example. 
The process involves two distinct matrices, A and $\{B_1, B_2\}$, for normalizing affinity values. 
Within matrix A, the element \(a_{ij}\) represents the relative importance of each pairwise criterion contrast in the criterion layer. 
The scaling method for \(a_{ij}\) is rule-based.
In this example, matrix A is straightforward, consisting of two rows and two columns. 
\(C_1\) comes from t-SNE method and \(C_2\) comes from the intersection on the Field Of View (FOV). 
The example assigns \(a_{12} = 7\) and \(a_{21} = \frac{1}{7}\).
}

\[
A=\begin{array}{c}
\quad\quad{C_1} \quad {C_2} \\
\begin{array}{c}
{C_1} \\
{C_2} \\
\end{array}
\left[\begin{array}{cccc}
1 & 7 \\
\frac{1}{7} & 1 \\
\end{array}\right]
\end{array}
\]

\begin{figure*}[h]
    \centering 
    % \vspace{-3mm}
    \subfloat[AdaFlow's fixed affinity matrix]{\label{matrix1}
    \includegraphics[height=0.26\linewidth]{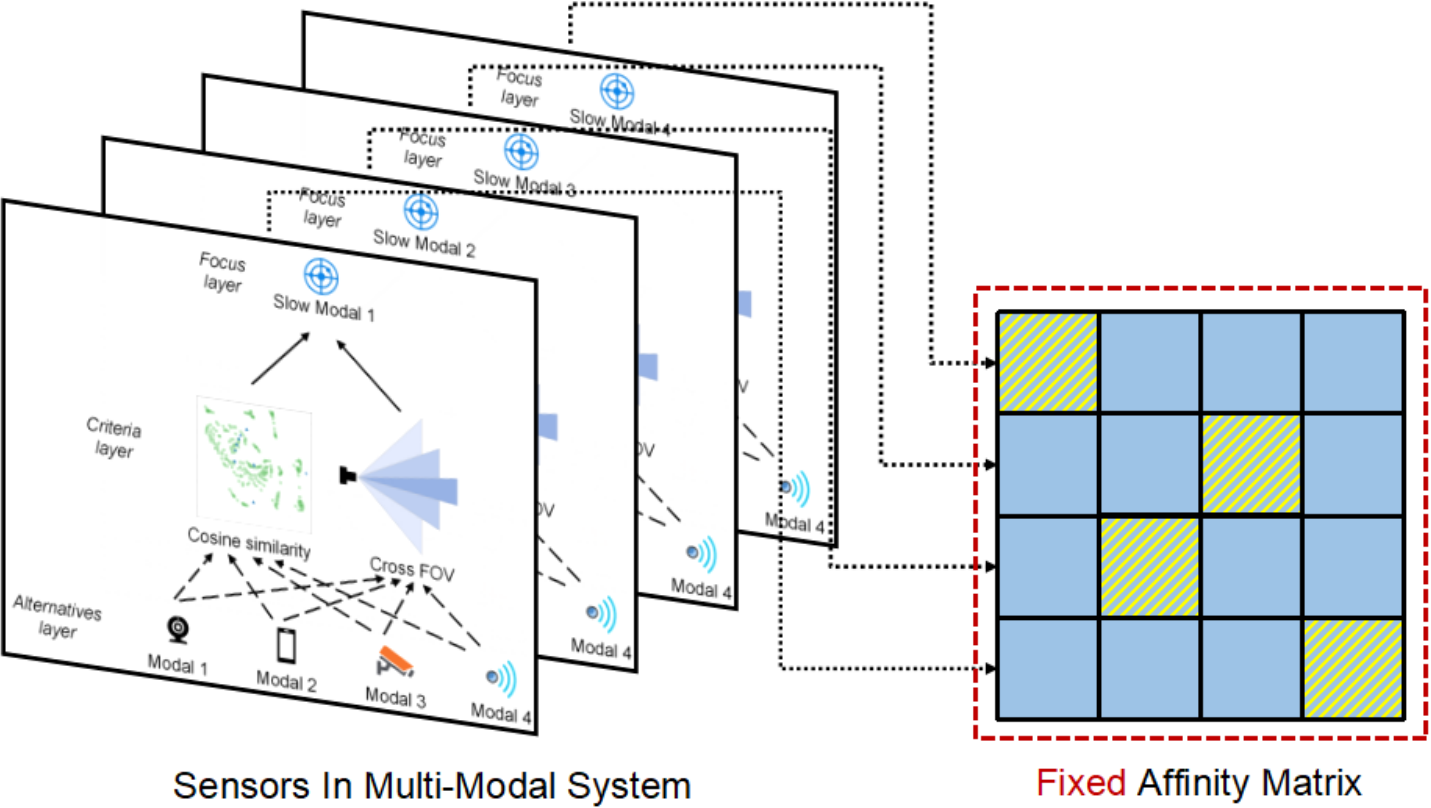}}
    \hspace{5mm}
     \subfloat[\sysnameposs modality-scalable affinity matrix]{\label{matrix2}
    \includegraphics[height=0.26\linewidth]{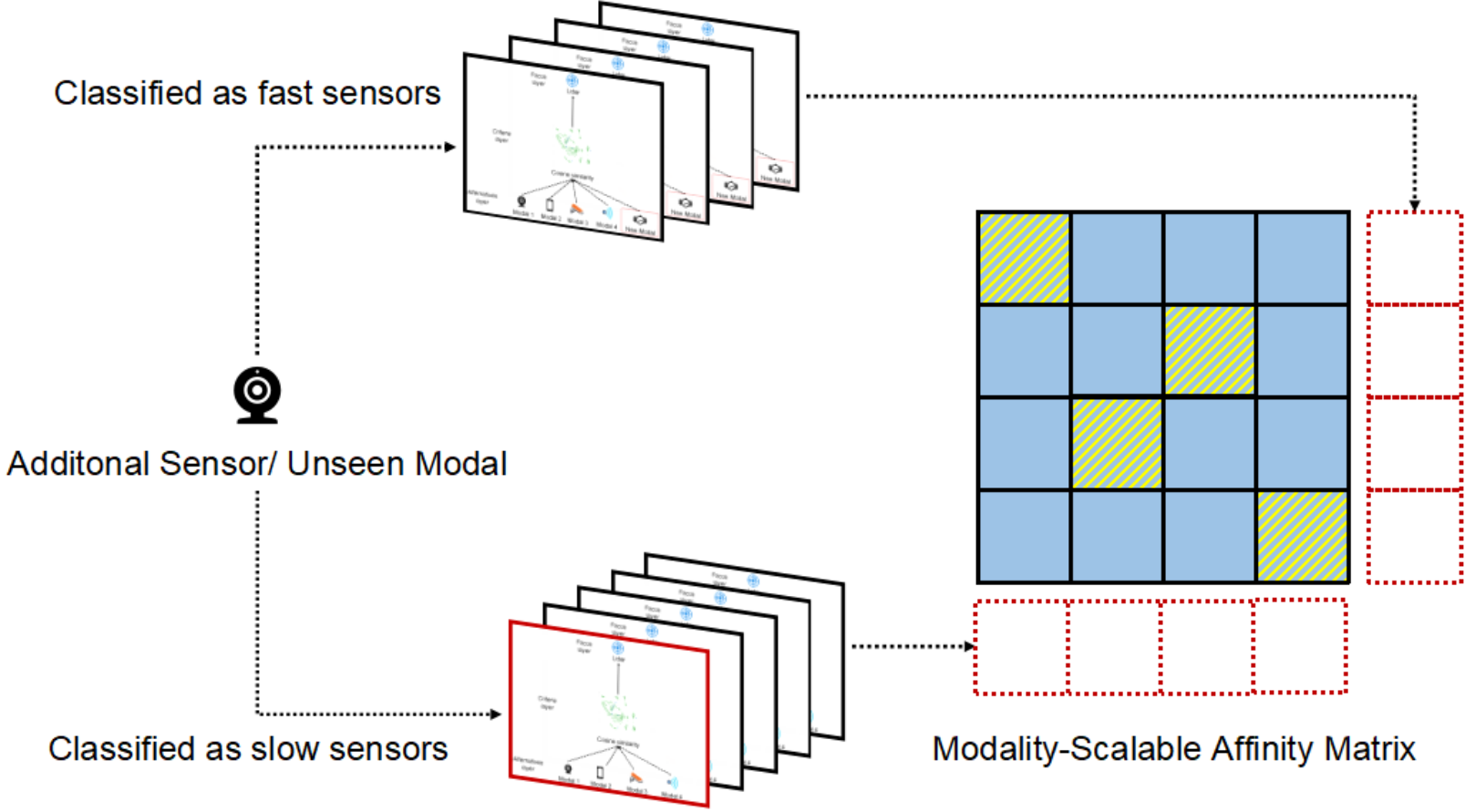}}
    %\vspace{-3mm}
    \caption{\wfm{(a) AdaFlow's fixed affinity matrix, (b) \sysnameposs modality-scalable affinity matrix.}}
    % \vspace{-2mm}
\label{figmatix}
\end{figure*}

\wyd{
In the matrix \(B_i\), the element \(b_{ij}\) represents the importance of sensor \(i\) relative to sensor \(j\) under criterion \(C_i\). 
This importance is derived by standardizing the average cosine or FOV values of each sensor in t-SNE. 
For example, with one Focus sensor and three Alternative sensors \(\{sensor_1, sensor_2, sensor_3\}\), the matrix \(B_i\) is a $3\times 3$ matrix.
}

\[
B_1=\begin{array}{c}
\begin{array}{c}
\end{array}
\left[\begin{array}{cccc}
1 & 2 & 5 \\
\frac{1}{2} & 1 & 2\\
\frac{1}{5} & \frac{1}{2} & 1\\
\end{array}\right]
\end{array}
B_2=\begin{array}{c}
\begin{array}{c}
\end{array}
\left[\begin{array}{cccc}
1 & \frac{1}{3} & \frac{1}{8} \\
3 & 1 & \frac{1}{3}\\
8 & 3 & 1\\
\end{array}\right]
\end{array}
\]
\wyd{
where \(A\) and \(\{B_1, B_2\}\) all pass the consistency test, their eigenvalues can be approximated by the arithmetic mean of their column vectors, thus obtaining the eigenvectors.
}

\[
W_1=\begin{array}{c}
\begin{array}{c}
\end{array}
\left[\begin{array}{cccc}
0.875  \\
0.125 \\
\end{array}\right]
\end{array}
W_2=\begin{array}{c}
{B_1} \quad\quad {B_2} \\
\begin{array}{c}
\end{array}
\left[\begin{array}{cccc}
0.594 & 0.082 \\
0.277 & 0.236 \\
0.129 & 0.682 \\
\end{array}\right]
\end{array}
\]
Then,
\[
W=W_1\cdot W_2=\begin{array}{c}
\begin{array}{c}
\end{array}
\left[\begin{array}{cccc}
0.53  \\
0.272 \\
0.198 \\
\end{array}\right]
\end{array}
\]
\wyd{
According to values in \(W\), for the Focus sensor, \(B_1\) has the greatest affinity, followed by \(B_2\), with \(B_3\) being the least.
This process yields a vector \(W\) for each sensor. These vectors are assembled to form the final affinity matrix, as depicted in \figref{matrix1}, which is crucial for robust decision-making in dynamic environments.
The outcome of this two-step strategy is a generalized affinity matrix. 
Please refer to~\cite{bib:adaflow:wu} for more detailed formula derivation and theoretical proof.
}

%子集和扩展

% \textbf{Limitations in xxx Senarios}.

\subsection{Prior-free Modality Affinity Refinement}
\textbf{Limitations of AdaFlow's Affinity Matrix}. 
AdaFlow's affinity matrix faces significant challenges in diverse mobile scenarios where new sensors without prior knowledge are integrated to enhance the multi-modal system's perception capabilities or refine local perception areas. 
This limitation primarily stems from the inherent design of the affinity matrix, which, as detailed in~\secref{paam}, relies on dimensionality reduction techniques to establish affinities between sensors and their fields of view (FOV). 
Among them, FOV presuppose the availability of prior knowledge, thus constraining the matrix's functionality when encountering \textit{unseen} modalities.

Furthermore, as illustrated in \figref{matrix1}, the structure of AdaFlow's affinity matrix is fixed upon construction. Introducing new sensors into this established matrix can lead to passive deformations, compromising the system's stability. 
To address these challenges, a scalable approach is desired to allow the affinity matrix to actively transform in response to the addition of new sensors, thereby maintaining stability and ensuring continuous system adaptability.

\textbf{Design}. 
To effectively manage \textit{unseen} modalities in our design, we propose a \textit{prior-free analysis} based on AdaFlow's affinity matrix.
Specifically, in scenarios where sensors need to be \textit{added}, we only utilize t-SNE exclusively as the criterion. This method does not strictly require historical data but instead focuses on the features of current sensor data.
This simplification allows us to bypass the necessity for prior knowledge, particularly concerning the field of view (FOV), which  plays a minor, auxiliary role in AdaFlow's affinity matrix.
By removing this component, we accept a trade-off in output accuracy for greater flexibility in integrating new sensor types without compromising the overall system functionality.

In our preliminary design, AdaFlow's affinity matrix categorizes sensor modalities into two groups: fast and slow, based on their asynchronous data arrival times, which are influenced by factors such as data volume and network bandwidth. 
This categorization also aids in reducing computational complexity during affinity refinement.
Building upon this foundation, we introduce the structured refinement strategy to the modality affinity matrix. 
Specifically, rows correspond to slow sensors, while columns represent fast sensors. 
This arrangement ensures matrix stability and responsiveness to dynamic changes in the sensor landscape. 
It enables seamless adaptation to new sensor modalities without incurring \textit{manual complexity} or significant computational overhead, making the approach both scalable and efficient.

\noindent$\bullet$ \textbf{Slow Sensors}.
The integration of a newly classified slow sensor into \sysnameposs modality affinity matrix is visually represented by the addition of a new row, as illustrated in~\figref{matrix2}. This addition reflects the affinity vector corresponding to the slow sensor. Given that each row in the matrix represents independent affinity vectors, as illustrated in \figref{matrix1}, the introduction of a new sensor does not necessitate recalculating existing vectors, thereby preserving computational resources.
In our \textit{prior-free analysis}, the matrix groups $\{B_1, B_2\}$ are simplified into a single matrix ${B_1}^{'}$. This new matrix maintains its original structure but is now populated with standardized affinity scores between the existing slow sensors and the newly added one. 
Meanwhile, matrix A is simplified to a unit matrix, this refinement reducing complexity and focusing the affinity analysis on the new interactions.
As a result, the affinity vector for the new sensor, $W={W_2}^{'}$, is directly derived from the eigenvectors of ${B_1}^{'}$, and \sysnameposs modality affinity matrix will formed as below:
\[
\begin{pmatrix}
 {W^1}^{T}\\
 \vdots \\
 {W^i}^{T}\\
{W^{i+1}}^{T}
\end{pmatrix}
\]

\noindent$\bullet$ \textbf{Fast Sensors}.
The integration of a newly classified fast sensor into the 
modality affinity matrix \sysnameposs is visually represented by the addition of a new column, as illustrated in \figref{matrix2}.
This process contrasts with the integration of slow sensors, which only requires adding a new row without necessitating recalculations of existing vectors.
However, the introduction of fast sensors entails a more complex adjustment. These sensors influence every affinity vector within the matrix. Consequently, matrix ${B_1}^{'}$, which is a simplified form  matrix groups $\{B_1, B_2\}$, undergoes a transformation to accommodate the new sensor dynamics. This necessitates recalculating the affinity vectors to reflect the updated sensor interactions, ensuring that the matrix accurately represents the enhanced sensing capabilities and maintains system integrity.
${B_1}^{'}$ will transformed as below:
\[
{B_1}^{'}=\begin{array}{c}
\begin{array}{c}
\end{array}
\left[\begin{array}{cccc}
\Huge B_1  & _{ \vdots }^ {\Huge b_{1j}} \\
 b_{i1}  \small \cdots  &1\
\end{array}\right]
\end{array}
\]
The rest of the calculation are the same as for \textit{slow sensors}, and \sysnameposs modality affinity matrix will formed as below:
\[
\begin{pmatrix}
 {{W^1}^{'}}^{T}\\
 \vdots \\
 {{W^i}^{'}}^{T}
\end{pmatrix}
\]

This refinement ensures that the matrix remains adaptable and efficient, capable of scaling to accommodate new sensors without significant recalculations or overhead.

%% file: body/inference.tex
\section{Lightweight Non-blocking Asynchronous Inference}
\label{sec:inference}

% This module aims to achieve adaptive imputation for asynchronous heterogeneous modalities to ensure non-blocking data stream without compromising accuracy.
% The system defines the problem as follows: a set of heterogeneous modals M = ($m_1$,$m_2$, ..., $m_n$), where M has two subsets, S and T, and M = S $\cap$ T. S represents the set of slow data streams, while T represents the set of fast data stream. For each data stream $s_i$ $\in$ S,$t_i$ $\in$ T, $a_i$ represents the transformation time from $s_i$ to $t_i$. Based on the arrival status of data streams and the corruption situation at each moment, the system adaptively determines which tasks are considered as source tasks S and which are target tasks T using the affinity matrix, in order to minimize the imputation time.

This subsection then explores the adaptive fusion of asynchronous and heterogeneous input via imputation, ensuring non-blocking inference without compromising accuracy and maintaining compute consumption low.
In a preliminary version of \sysname, we used a CNN-based GAN for one-fit-all data imputation as in AdaFlow~\cite{bib:adaflow:wu}. 
However, for complex data like point clouds, CNN-based GANs require deeper architectures and larger parameters to capture data correlations, which can hinder real-time performance on resource-limited mobile platforms. 
In contrast, the Transformer’s scalability and parallel processing capabilities make it a more efficient and lightweight alternative for \sysname, enabling one-fit-all data imputation without sacrificing performance.
% \TODO{also highlight difference to AdaFlow, and Why.}

% As shown in \figref{partial}.

% As shown in \figref{partial}. 
% An example from non-top LiDAR, sensor B, is selected. 
% Due to the different installation positions and sensing view from senor A, the affinity criteria for selecting a paired sensor should focus more on consistency rather than affinity. 
% So fast data with larger cosine similarities are selected for fusion based on the affinity matrix. 
% The selection criteria are that larger cosine similarities, indicative of smaller average cosine distances, enhance the matching process, as these data modes exhibit high similarity. 
% Non-top LiDAR sensors, which generate 120° point clouds, benefit from selecting highly consistent RGB images to restore spatial information in significantly missing sections.

\subsection{Lightweight One-fit-all Data Imputation}
\label{gan}
\textbf{Why Transformer}. The Transformer architecture has strong applicability across diverse \textit{data types} and \textit{tasks}, showing particular promise for data imputation for three key reasons:
\textit{i)} The self-attention mechanism enables Transformers to effectively capture long-range dependencies in data streams, making them well-suited for mobile data imputation~\cite{bib:trans:v}.
\textit{ii)} Transformers can be scaled (in terms of layers, heads, \textit{etc.}) to suit varying task complexities, offering flexibility for optimizing dynamic data imputation tasks~\cite{bib:bert:d}.
\textit{iii)} The architecture allows for parallel processing of entire sequences, improving efficiency, critical for real-time performance demands on resource-constrained mobile devices with complex input~\cite{bib:lora:h}.

\textbf{Limitations of Prior Arts}.
While directly using  Vanilla Transformer~\cite{bib:trans:v} poses challenges in parameter storage, we adopt the Swin Transformer~\cite{bib:swin:l} for its computational efficiency. 
The Swin Transformer reduces complexity by restricting self-attention calculations to localized windows, which also improves scalability. 
Additionally, its \textit{Shifted Window} mechanism enhances inter-window communication, boosting the model’s expressive power. 
This localized attention strategy is particularly well-suited for imputation tasks, making the Swin Transformer a highly effective choice for handling such generalized and dynamic data imputation requirements on mobile devices.
However, due to \textit{dynamic asynchrony} and \textit{heterogeneity} in mobile sensing patterns, directly applying Swin Transformer to a distributed multi-modal system is often ineffective. 
This is because the traditional Swin Transformer typically assumes \textit{synchronous} inputs with \textit{fixed} input volume and type, which does not always hold in real-world mobile distributed systems.
For example, slow LiDAR data with varying sensing patterns requires video data from different cameras for imputation.

\begin{figure}[t]
	\centering
	\includegraphics[width=0.98\linewidth]{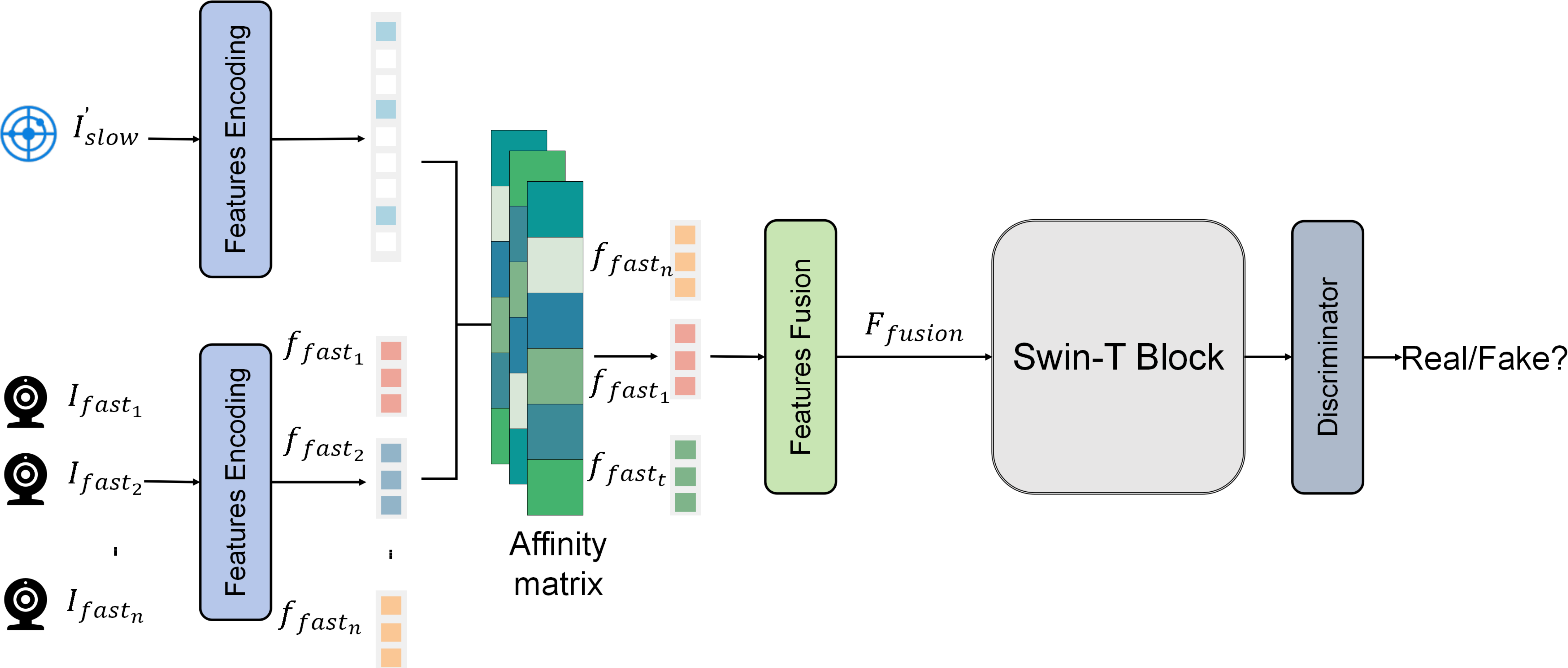}
        % \vspace{-3mm}
	\caption{Illustration of multi-modal lightweight Swin Transformer (MMLST) for real-time data imputation on resource-limited mobile devices. }
        % \vspace{-3mm}
	\label{fig:tgan}
\end{figure}

%总体目标
%如图8所示，我们的模型包括了几个模块，和输入、输出的关系. 如图9所示，xxx、xxx、xxx、xxx多少个stage（名字）--- 总述之间关系。
%从前到后，一个模块一个模块的去讲。每个模块都要先说问题、挑战、已有不足（existing  methods leverage xxx or xxxx），以及你的设计。
\textbf{Our Scope}.
To tackle these challenges, we extend the Swin Transformer to multi-modal settings and integrate it with the proposed affinity matrix using a lightweight architecture, namely the Multi-modal Lightweight Swin Transformer (MMLST). 
The Swin Transformer captures correlations between fast and slow data modalities (\eg RGB images and point clouds) by utilizing attention windows of varying sizes. 
The architecture of the MMLST is shown in \figref{fig:tgan}. 
The MMLST architecture has two key components: data resampling with an affinity matrix and data imputation using the lightweight Swin Transformer (Swin-T) block.
In the resampling part, the affinity matrix (as detailed in \secref{matrix}) enables the adaptive selection and fusion of asynchronous, heterogeneous inputs. 
After resampling the fast data streams, the remaining data is fused with the slow stream and passed to the Swin-T block for further processing.

\begin{figure*}[h]
	\centering
	\includegraphics[width=0.7\linewidth]{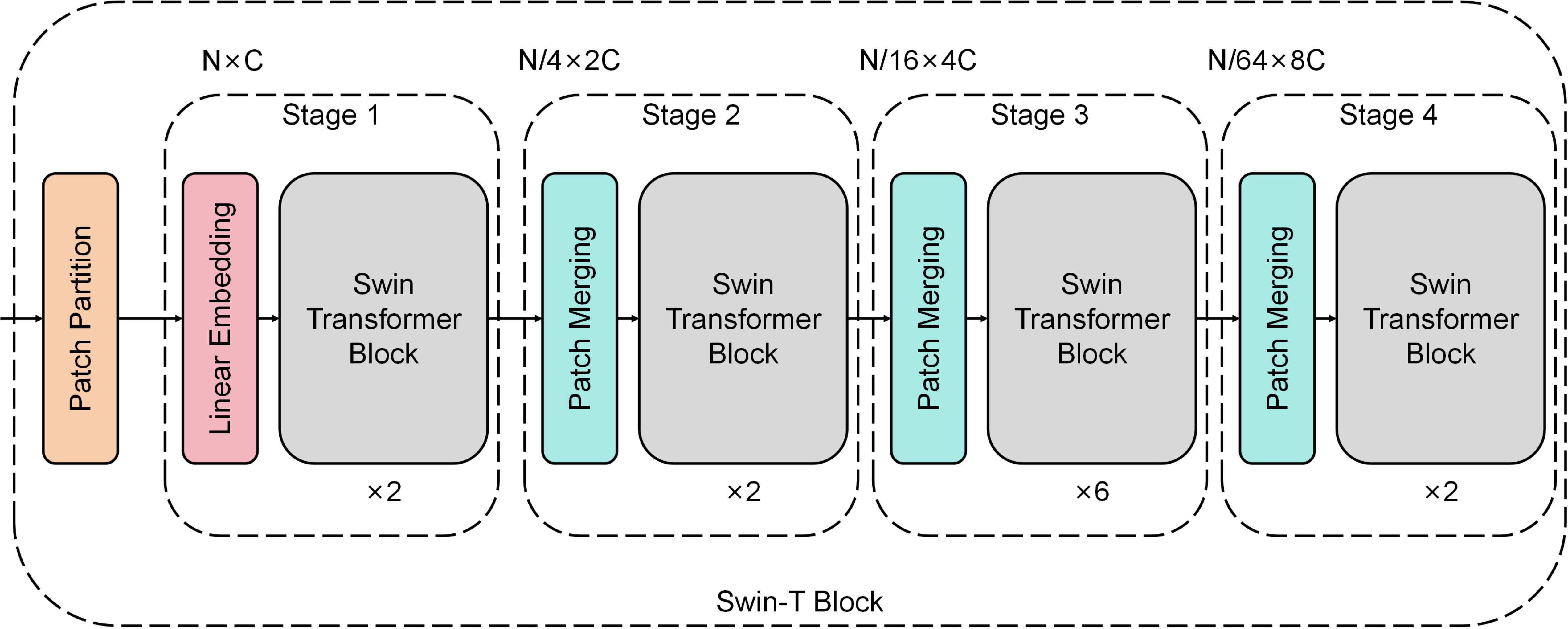}
        % \vspace{-3mm}
	\caption{Illustration of the Swin-T block in the proposed MMLST.}
        % \vspace{-3mm}
	\label{fig:st_a}
\end{figure*}

\figref{fig:st_a} illustrates the Swin-T block, as shown in \figref{fig:st_a}, which consists of four stages.
Initially, the fused features are segmented into equal patches by a patch-splitting module, allowing the model to learn the characteristics of missing data locally, particularly suited to data imputation tasks. 
To better handle multi-modal inputs, we redefine the traditional $Height \times Width$ or $Height \times Width \times Depth$ configurations. Instead, we specify the number of patches, 
$N$, based on token count rather than shape, improving the model's adaptability to diverse input modalities.
To enhance the Swin-T block's ability to learn inter-modal relationships, we further introduce a linear embedding layer that projects features into a higher-dimensional space, denoted by 
$C$. 
Finally, Swin Transformer blocks are applied to these patch tokens, maintaining the token count 
$N$ throughout Stage 1, which also includes the linear embedding.

\begin{figure}[t]
	\centering
	\includegraphics[width=0.9\linewidth]{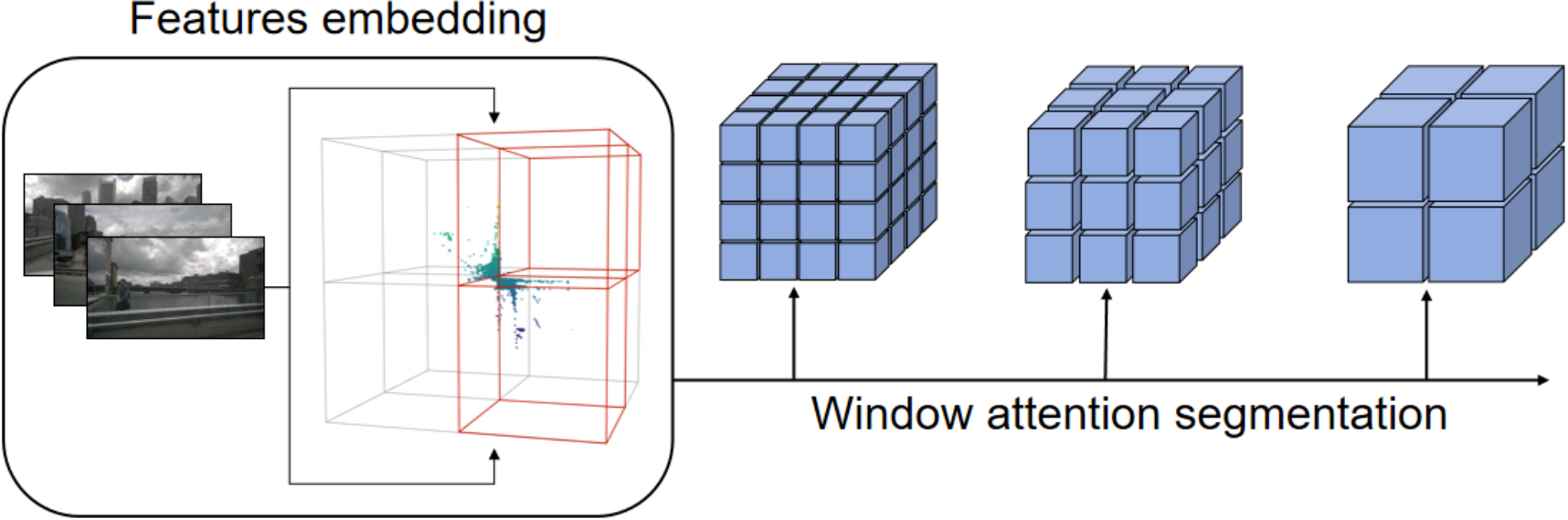}
        % \vspace{-3mm}
	\caption{Shifted Window applied in RGB images and cloud point.}
        % \vspace{-3mm}
	\label{fig:st}
\end{figure}

As the network depth increases, the number of tokens decreases due to patch merging layers. The initial patch merging layer aggregates features from four neighboring patches. Subsequent Swin-T blocks transform features while maintaining a resolution of \(N/4\) and doubling the output dimension to \(2C\). 
This process, consisting of the patch merging and feature transformation, forms Stage 2.
Stages 3 and 4 follow the same procedure, reducing the output resolutions to \(N/16\) and \(N/64\), respectively. 
Compared to the unoptimized Swin Transformer, we accelerate the patch merging process and reduce the number of stages from \textit{many} to just \textit{four}. 
This optimization significantly lowers the number of parameters, making the model more efficient for resource-constrained environments. Additionally, we maintain the basic setup of the two successive Swin Transformer blocks.

Notably, we \textit{observe} that the shifted window mechanism of the Swin Transformer can seamlessly integrate with the proposed \textit{affinity matrix}. 
Specifically, after the patch partitioning, the features from fast data and the corresponding features from slow data that require imputation are localized within specific patches. 
This strategic localization enhances the model's ability to effectively capture correlations between modalities, thereby improving the accuracy and efficiency of the feature fusion process in resource-constrained environments.

\textbf{Training of MMLST}.
\rev{The training loss function consists of two components: the Swin-T loss \(\mathcal{L}_{Swin}\), and a loss function for slow data stream characteristics \(\mathcal{L}_{*}\), which measures the discrepancy between modalities.
Inputs to the Swin-T block \(S\) include complete fast data stream \{\textbf{\(I_{fast_1}\)}, \textbf{\(I_{fast_2}\)}, ..., \textbf{\(I_{fast_n}\)}\} and the incomplete slow data stream \textbf{\(I_{slow}^{'}\)}.
This setup enables \(S\) to generate the synthetic slow data \textbf{\(I_{slow}^{*}\)}. 
The loss function is formulated as follows:}
\begin{equation}
\begin{split}
&\{\textbf{$F_{fast_1}$},\textbf{$F_{fast_2}$},...,\textbf{$F_{fast_n}$}\}=\\&\textbf{$f_{fast}$}(\{\textbf{$I_{fast_1}$},\textbf{$I_{fast_2}$},...,\textbf{$I_{fastn}$}\})
\end{split}
\end{equation}
\begin{equation}
\textbf{$F_{slow}$}=\textbf{$f_{slowR}$}(\textbf{$I_{slow}^{'}$})
\end{equation}
\begin{equation}
\textbf{$F_{fusion}$}=
g(h(\{\textbf{$F_{fast_1}$},...,\textbf{$F_{fast_n}$}\}),\textbf{$F_{slow}$})
\end{equation}
\begin{equation}
\textbf{$I_{slow}^{*}$}=S(\textbf{$F_{fusion}$})
\end{equation}
\begin{equation}
\begin{split}
&   
\mathcal{L}_{Swin}=\mathbb{E}_{\textbf{$I_{slow}$}}[logD(\textbf{$I_{slow}$})]\\
&+\mathbb{E}_{\textbf{$I_{slow}^{*}$}}[log(1-D(\textbf{$I_{slow}^{*}$}))]
\end{split}
\end{equation}
where the \(S\) seeks to minimize Equ. 10, while the discriminator \(D\) strives to maximize it. 

% \wfm{\textbf{Comment: the authors use chamfer distance loss to train ACGAN, which seems to be mainly applicable to point cloud data. Does this mean that if one wants to generate data of different modalities, the system needs to modify the loss function and retrain ACGAN?}}

For example, in the task of 3D object detection for autonomous driving assistance, the camera data is always the fast stream, whereas the LiDAR data is slow. 
Due to the characteristic nature of point cloud produced by LiDAR, we opt for the Chamfer distance loss \(\mathcal{L}_{CD}\)(Chamfer distance loss) between real point cloud \(X\) and synthetic point cloud \(Y\) as \(\mathcal{L}_{*}\) which is expressed as follows:
\begin{equation}
X=\{x_i\}_{i=1}^N Y=\{y_i\}_{i=1}^M,X=\textbf{$I_{LiDAR}^{*}$} 
 Y=\textbf{$I_{LiDAR}$}
\end{equation}
\begin{equation}
\mathcal{L}_{CD}=\sum_{x \in X} \min_{y \in Y} \|x - y\|_2^2 + \sum_{y \in Y} \min_{x \in X} \|y - x\|_2^2
\end{equation}
For each point \(x\) in point cloud \(X\), we identify the nearest point \(y\) in point cloud \(Y\) and compute the squared Euclidean distance between them, summing these values for all points in $X$. 
Similarly, for each point \(y\) in \(Y\), we find the nearest point \(x\) in  \(X\), calculate the squared Euclidean distance between them, and sum these values for all points in $Y$.
Thus, our final objective function becomes:
\begin{equation}
\mathcal{L}=arg\,\underset{S}{\min} \, \underset{D}{\max}\,\mathcal{L}_{Swin}+\lambda_{CD}\mathcal{L}_{CD}
\end{equation}
where $\lambda_{CD}$ is a hyper-parameter to tune the data fitting and model complexity.

% As a separate note, in autonomous driving, LiDAR data is generally slower than camera data. For instance, any combination of cameras $I_{fast_1}$, $I_{fast_2}$, and $I_{fast_3}$ (with asynchrony) can generate LiDAR data $I_{LiDAR}'$, with the loss function incorporating point cloud. 
% If the downstream task is fixed, asynchronous data can fill in for imputing the slower LiDAR without retraining. 
% The loss function can also be adjusted for other modalities in different tasks.

\subsection{Affinity-aware Sub-graph Selection}
\label{graph}
% 现有方法assume互补性一致性是不变的，however， 对不同的sensor deploy pattern，互补性一致性会发生改变。
\rev{Existing data imputation methods, such as those discussed in~\cite{bib:gain:y,bib:miwae:m}, typically presuppose a fixed set of modalities for data fusion and imputation, assuming static consistency and complementarity based on predetermined sensing patterns.
However, they often fail in dynamic mobile environments, which are also characterized by resource constraints~\cite{bib:lls:L}.}
Studies~\cite{bib:incomplete:w,bib:less:d} reveal potential drawbacks. Specifically, \cite{bib:incomplete:w} shows that adding more modalities without \textit{key} ones can introduce noise and redundancy, degrading performance. 
Similarly, \cite{bib:less:d} reports that not all modalities contribute positively, with some introducing unnecessary noise. 
\wyd{
To address this, \sysname applies \textit{affinity-aware sub-graph selection} to selectively choose modalities that enhance fusion for better imputation (as we will show in \secref{val}).
}

For instance, in the real-world deployment of \sysname in autonomous driving assistance systems, which integrates two LiDAR sensors and five cameras, we demonstrate the effectiveness of our approach under practical, resource-constrained conditions.
The LiDARs are strategically positioned at the top and front, while the cameras capture images from various perspectives, including front, front-left, left, front-right, and right. 
These sensors correspond to sensors A$\sim$G in \figref{full} and \figref{partial}.
Specifically, for the top LiDAR, which produces 360° point clouds in the slower data stream, we set a missing rate of 75\% to mimic the delay typically observed in slow data streams, as shown in \figref{full}. 
Given the broad sensing view of sensor A, its criteria for selecting fusion modalities should prioritize \textit{complementarity} over \textit{consistency}. 
Thus, fast data in sensor A with smaller cosine similarity are preferred for fusion to match the slow data stream, as detailed in \secref{full}. 
Here, smaller cosine similarity, indicating larger average cosine distances, denotes a high level of \textit{complementarity} between heterogeneous modalities. 
With a 75\% missing rate of front LiDAR sensor B, selecting RGB images from other sensors with higher cosine similarity provides limited benefits in fusion accuracy.
Similarly, for point clouds generated by other LiDAR sensors in the slow stream, we also maintain a 75\% missing rate.

\wyd{We note that the imputation may \textit{vary} according to different data arrival patterns. 
\textit{Affinity-aware sub-graph selection} can be described as follow}: given a set of heterogeneous modalities \( M = \{m_1, m_2, \ldots, m_N\} \) as input, 
output the map-wise complementarity or consistency decision $m_{fi}\rightarrow m_{sj}$, where $m_{sj}$ is the slow modality and $m_{fi}$ is the fast modality. It can be formulated as $m_{fi}\rightarrow m_{sj}$ = $\text{F}(M)$.
For $\text{F}(M)$, we assess complementarity and consistency using the intersection over union between modalities. 
The intersection over union between modalities is defined as:
% First of all to distinguish criteria of affinity of each \(m_{sj} \in S\). We shall have intersection of the FOV between \(m_{fi}\) and \(m_{sj}\):
\begin{equation}
    v_{ij} =  \frac{\text{FOV}_{m_{sj}} \cap \text{FOV}_{m_{fi}}}{\text{FOV}_{m_{sj}} \cup \text{FOV}_{m_{fi}} }
\end{equation}

And if \(v_{ij}\) $\not=$ 0, \(v_{ij} \in V_j\). For each \(m_{sj} \in S\), we set:
\begin{equation}
 m_{fi}\rightarrow m_{sj} = 
 \begin{cases}
Consistency\ ,  &\exists v_{ij}=0\\
Complementarity, &\exists\mkern-10.5mu/ \ v_{ij}=0
\end{cases}
\end{equation}

For \(m_{sj}\), its' criteria of affinity has settle down, \( a_{ij} \) is the affinity between \( m_{fi} \) and \( m_{sj} \),
Then we have:
\begin{equation}
    A=\sum_{}^{j} \sum_{i=1}^{k}a_{ij}
\end{equation}
Where the value of \(k\) as shown below:
\begin{equation}
k=\lfloor |V_j| \times r \rfloor
\end{equation}

Where \(r\) is the data missing rate of partial (slow) data and $|V_j|$ is the number of intersecting FOV sensors. We need to discuss for value \(k\):
\begin{equation}
k=
\begin{cases}
k, k \not= 0 \\
1, k = 0 \ |V_j| \not= 0\\
0, k=0  \ |V_j| =0
\end{cases}
\end{equation}
By maximizing \(A\) (Equ. 16), we obtain the 
$m_{fi}\rightarrow m_{sj}$ map. 
% Then by maximizing \(A\), we can have map \( \{m_{fi},m_{fi+1},\ldots,m_{fi+k-1}\}\rightarrow m_{sj}\).

Based on the arrival speed and availability of input data streams, \sysname adaptively determines the selection of \textit{affinity sub-graph} for data fusion by maximizing \(A\), \ie which modality and modality \( S \) to fuse, and which are the target tasks \( F\) using the selected affinity sub-graph, to maximize the imputation effect.
In particular, based on the affinity matrix as discussed in \secref{matrix}, the sub-graph selection criteria for runtime fusion are dynamically determined by the specific requirements of the downstream task, influenced by the asynchrony and heterogeneity of input data. 
% \wfm{As illustrated in the case, we specify the selection criteria for affinity according to the specific requirements of the downstream task, and then use the affinity matrix to analyze each \( s_i \in S \) to obtain the global \textit{affinity sub-graph}.}

%% file: body/experiments.tex
\section{Experiments}
\label{exp}
\subsection{Setups}

\textbf{Implementation.} 
\sysname is implemented using Python 3.9 and PyTorch 1.11 on a server equipped with an RTX3090 GPU, Intel(R) Xeon(R) Gold 6133 CPU @ 2.50GHz, and 256GB RAM. 
For distributed multi-modal data on mobile devices, we realize both simulation with real-world datasets and actual collections. 
We also simulated the bandwidth of the real car Internet, which is 100Mbps.
In the implementation of the affinity matrix (\secref{matrix} ), we designate LiDAR as the primary modality and cameras as secondary modalities. For the MMLST architecture (\secref{gan}), we employ two feature-capturing networks to process the raw data. 
These networks, via the affinity matrix, select the appropriate modalities for fusion. 
The resultant fusion features are then inputted into the Swin-T block to synthesize the primary modality data.

\textbf{Tasks, Datasets, and Model.} We experiment with 4 real-world distributed multi-modal applications.
They are widely used to evaluate multi-modal fusion performance~\cite{bib:nuscenses:ca,bib:kitti:g,bib:da:m}.
\begin{itemize}
\item \textbf{BEVFusion ($T_{1}$)} is a real-world multi-modal 3D object recognition task using a bird's-eye view.
The \textbf{dataset} (nuScenes ~\cite{bib:nuscenses:ca}) contains 2000 training frames and 81 testing frames, each featuring 6 camera views (fast data) and 1 LiDAR views (slow data).
The \textbf{model} for BEVFusion extracts a bird's-eye view (BEV) from both the camera and LiDAR branches.
% 车载物联网，每秒100字节。数据量比较小。50毫秒。用的dataset，有慢速和快速数据之分吗？点云数据，至少要大两倍。目的是告诉读者，这是代表性的数据集，要反应场景下的真实数据流分布。

\item \textbf{PointPillars ($T_{2}$)} is another real-world 3D object recognition model. 
The \textbf{dataset} from KITTI~\cite{bib:kitti:g} includes 2500 frames for training, and 81 frames for testing, all collected from actual autonomous driving systems, with each frame featuring 4 camera views (fast data) and 1 LiDAR view (slow data)
The \textbf{model} projects both camera and LiDAR data into a unified space and merges multiple views.
\item 
\textbf{LLM-based driver behavior recognition ($T_{3}$)} employs \textbf{Large Language Model (LLM)}~\cite{bib:one:h} to categorize driver behaviors into 12 classes using the Drive\&Act dataset~\cite{bib:da:m}, which comprises 6 infrared camera views (fast data) and 1 standard camera view (slow data).
\item 
\textbf{LLM-based event recognition (\(T_4\))} employes \textbf{LLM} for audio-visual event recognition across 28 categories using the AVE dataset~\cite{bib:ave:t} which contains audio (fast data) and video (slow data).

% 数据大小之比7:5，有快速和慢速。模拟车载以太网，100Mbps。调研车企数据得来的。

\end{itemize}

% 这个方法还挺有意思的。
% 为什么别人的方法（A+B，B+C），为什么别人方法不能轻易的generalize 到不同，应用scope非常窄，遇到不同的应用和输入情况他们必须重新设计，我们可以直接部署，Unique challenge。做了一些相关解释不够充分。
% 证明在不同模态间都可以用，实验要体现下来。 
% 主打泛化性，方法是足够novel。

\textbf{Baselines.} We adopt five a-/synchronous fusion methods as comparison baselines to validate the affinity matrix.
\begin{itemize}
\item \textbf{Synchronous fusion} (blocking mechanism, BM)~\cite{bib:patch:w}: 
the inference pipeline waits until all input data arrive. 
It provides an upper-bound accuracy.
\item \textbf{Asynchronous fusion} (non-blocking mechanism):
\begin{itemize}
    \item Sparse Point Cloud (SPC)~\cite{bib:mdl:n}: discards the slow data for inference which is the fastest method but exhibits the lowest accuracy.
    \item PCN~\cite{bib:pcn:Y}: is a DL model to impute 3D point clouds. When the point cloud is missing due to data asynchrony, we use it to complete the point cloud.
    \item KNN~\cite{bib:knn:c}: is a traditional imputation method, which classifies an input by majority class among its $K$ nearest neighbors to selectively discard data.
    \wyd{\item AdaFlow~\cite{bib:adaflow:wu}: has the same data resampling effect as \sysname under fixed modalities.
    We use it to test the Swin-T block's lightweight characteristics.}
    \item Only Swin-T block: has the same imputation network structure as \sysname, but selects fixed or random fast data to impute the slow data. 
    We use it to test the effectiveness of the affinity matrix.

\end{itemize}
% \item \textbf{OneLLM} is a Large Language Model (LLM) platform. 
% We employ OneLLM for driver behavior recognition across 12 categories using the Drive\&Act dataset, which includes two modalities: RGB and Infrared Spectroscopy (IR). 
% The RGB features a sensor with a right-middle perception angle, capturing comprehensive driver information. The IR modality comprises six sensors with varying angles: right-middle, front-top, right-top, back, face-view, and left-top. 
% Considering data size, RGB is set as slow data and IR as fast data. We test the following two methods:
% \begin{itemize}
%     \item Dynamic modal combination: for each RGB frame, OneLLM randomly selects one from six IRs.  
%     \item Fixed modal combination: for each RGB frame, OneLLM selects a pre-defined best one from six IRs.
% \end{itemize}
\end{itemize}

\subsection{Performance Comparison}
\label{performance}
% \subsubsection{Accuracy vs. Latency }
We evaluate the inference accuracy and latency of \sysname against five baseline methods (BM, SPC, PCN, KNN and AdaFlow) across BEVFusion ($T_1$) and PointPillars ($T_2$) tasks under various missing rates of slow data, which simulate data asynchrony. 
For example, a 25\% missing rate of slow data represents a delay of 26ms in dataset nuScenses with 100Mbps.

\begin{figure}[h]
  \centering
  %\vspace{-3mm}
  \subfloat[BEVFusion, missing 50\% LiDAR.]{
    \includegraphics[height=0.12\textwidth]{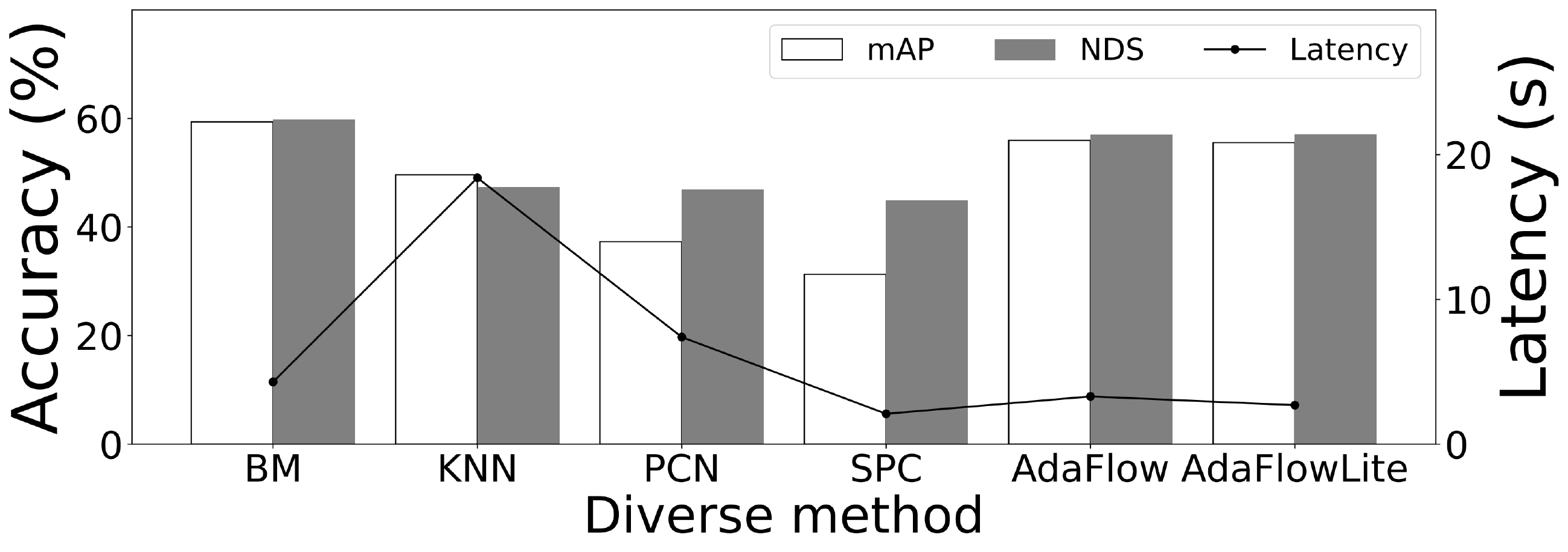}
    \label{fig:bev1}
  }
  \\
  \vspace{-3mm}
  \subfloat[BEVFusion, missing 75\%  LiDAR.]{
    \includegraphics[height=0.12\textwidth]{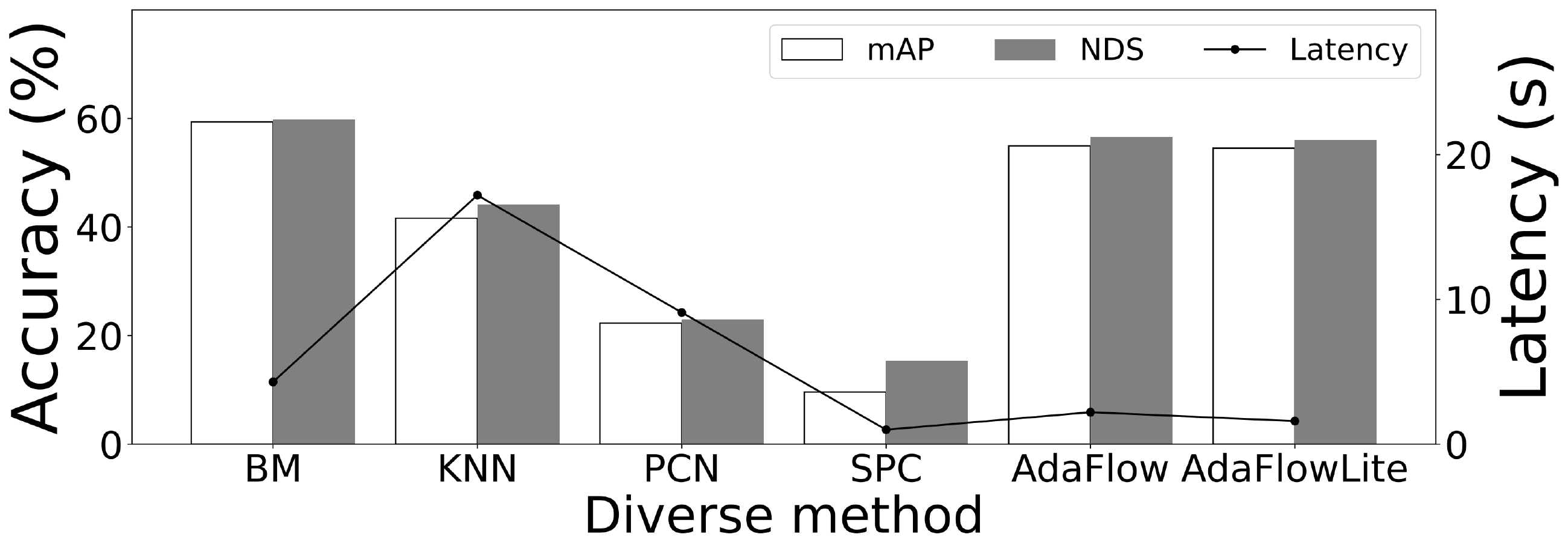}
    \vspace{-2mm}
    \label{fig:bev2}
    \vspace{-2mm}
  }
\\
\vspace{-3mm}
    \subfloat[PointPillars, missing 25\% LiDAR.]{
    \includegraphics[height=0.12\textwidth]{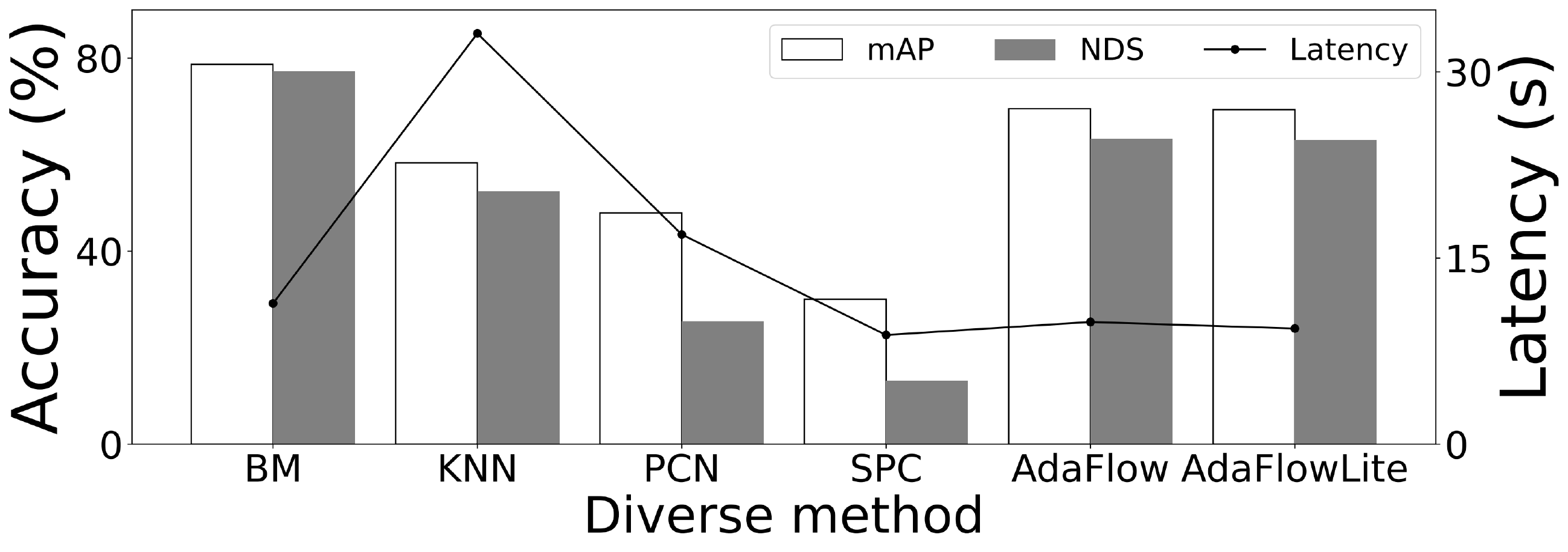}
    \label{fig:poin1}
    %\vspace{-10mm}
  }
\\
\vspace{-3mm}
  \subfloat[PointPillars, missing 50\% LiDAR.]{
    \includegraphics[height=0.12\textwidth]{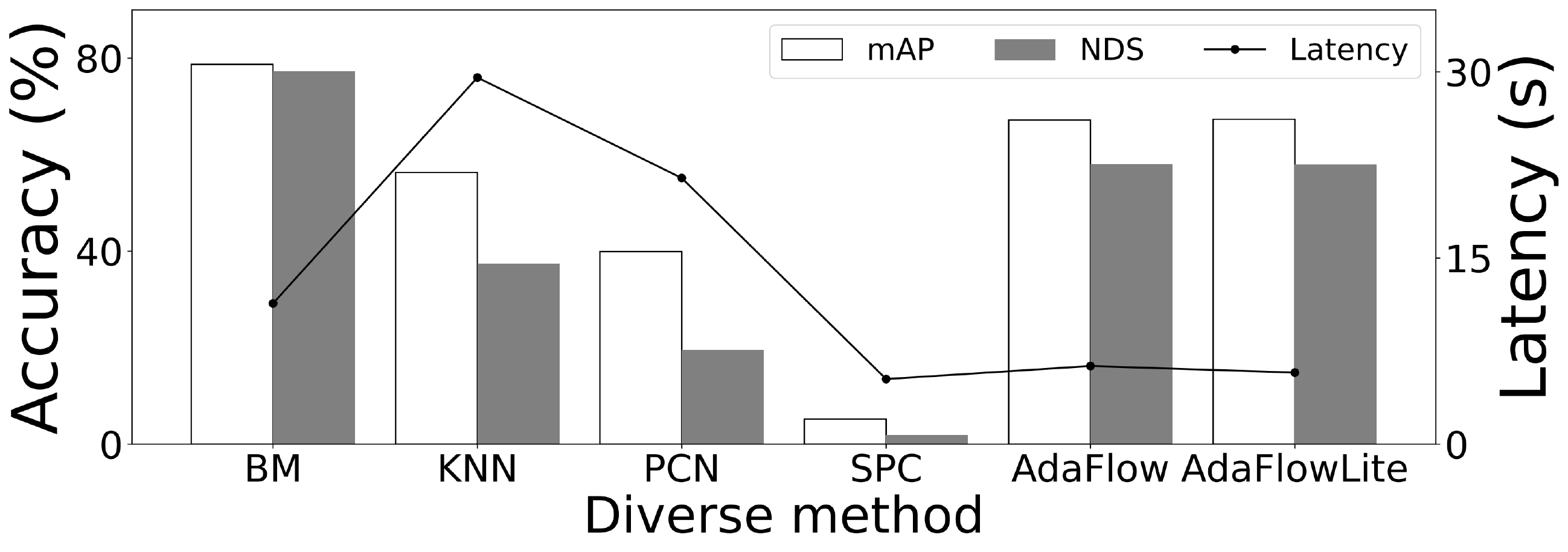}
    \label{fig:poin2}
  }
  \caption{
  Comparison of accuracy and inference latency between \sysname and baselines in two tasks, under varying missing rates of slow data (\ie lagging delay).}
  \vspace{-3mm}
  \label{fig:exp:bevpoin}
\end{figure}

\figref{fig:exp:bevpoin} shows the results.
\textit{First}, \sysname exhibits the best trade-off between inference accuracy and latency compared to the three asynchronous fusion methods (SPC, PCN, and KNN), and achieves almost as good accuracy as AdaFlow.
\textit{Second}, \sysname outperforms the three asynchronous fusion methods (SPC, PCN, and KNN) under various missing rates of slow data. 
For example, as shown in \figref{fig:bev2}, with 75\% missing rate in BEV, \sysname demonstrates accuracy improvements of 44.9\%, 32.2\%, 12.9\% compared to SPC, PCN, and KNN, respectively.
\textit{Third}, with acceptable accuracy, \sysname achieves the lowest latency compared to other baselines. 
As shown in \figref{fig:poin2}, with 50\% missing rate of slow LiDAR data in PointPillars, \sysname achieves 49.0\%, 80.4\%, 73.1\%, 8.3\% latency reduction compared to BM, KNN, PCN, and AdaFlow. 
Besides, \sysname improves 62.1\% accuracy compared to SPC with only 0.5s slower than SPC for 81 frames.

\textbf{Summary.} \sysname offers the optimal balance between accuracy and latency, making it a promising solution for latency-sensitive tasks that often involve asychronous data.

% \begin{figure}[h]
%   \centering

%   %\\
%   % \vspace{-1mm}
%   \caption{Comparison of accuracy vs. training time between \sysname and other baselines in PointPillars under diverse missing rate based on dataset's characteristics.}
%   \label{fig:exp:poin}
% \end{figure}

\subsection{Parameter Comparison}
We evaluate the numbers of parameters, FLOPs, and storage requirements of \sysname against AdaFlow. \tabref{table:para} shows the results. \sysname achieves a reduction of 49.5\%, 51.5\%, and 49.5\% in the number of parameters, FLOPs, and storage requirements, respectively, compared to AdaFlow. As discussed in \secref{performance}, \sysname maintains nearly identical accuracy with a 50\% reduction in these metrics, making it more suitable for resource-constrained environments.

\begin{table}[h]
\centering
\caption{Comparison of parameters between AdaFlow and \sysname.}
\label{table:para}
\small
\scalebox{0.8}{
\begin{tabular}{|c|c|c|c|}
\hline
Models & Parameters & FLOPs(GFLOPs) & Storage Requirements(MB) \\
\hline
AdaFlow & 354233600 & 4.648 & 1351.29 \\
\hline
\sysname & 178836346 & 2.256 & 682.21 \\ 
\hline
\end{tabular}}
\end{table}
\subsection{\sysnameposs Data Imputation Performance}
\begin{figure}[t]
	\centering
	\includegraphics[height=0.13\textwidth]{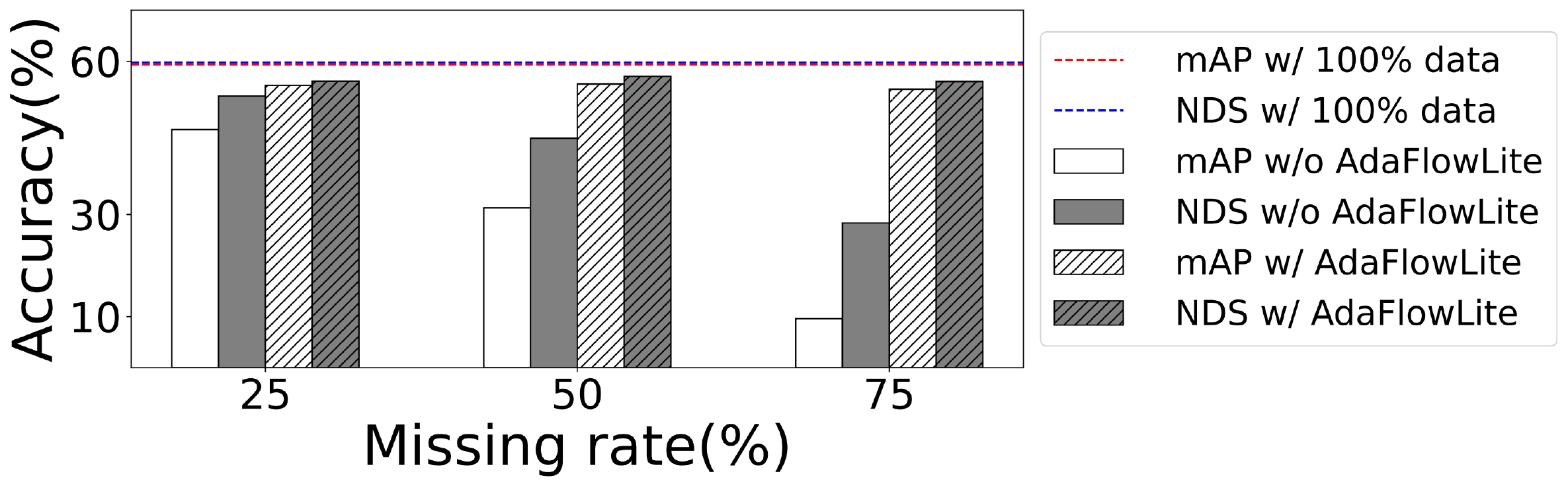}
	\caption{\sysnameposs imputation performance in BEVFusion under different missing rates of slow data.}
 \vspace{-3mm}
	\label{fig:BEV}
\end{figure}

\begin{figure}[t]
	\centering
	\includegraphics[height=0.12\textwidth]{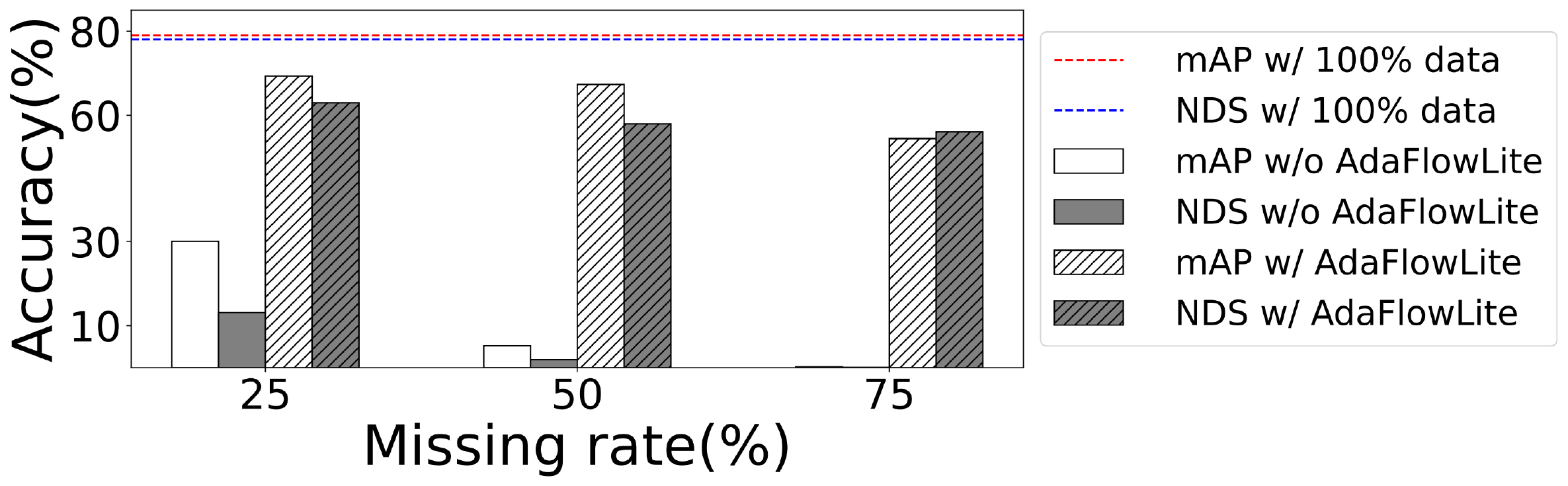}
         \vspace{-3mm}
	\caption{\sysnameposs imputation performance in PointPillars under different missing rates of slow data.}
 \vspace{-3mm}
	\label{fig:POIN}
\end{figure}

We evaluate \sysname's data imputation module on BEVFusion ($T_1$) and PointPillars ($T_2$) under different data missing rates of slow data.
% \figref{}\todo{ref} shows the results.
\textit{First}, \sysname efficiently prevents significant accuracy decline at varying slow data missing rates.
As shown in \figref{fig:BEV}, \sysname suffer only 4.1\%, 3.8\%, and 4.8\% accuracy decline with data missing rates at 25\%, 50\%, and 75\%.
While SPC incurs up to 12.7\%, 28.1\% and 49.8\% accuracy drop, respectively.
\textit{Second}, \sysname is more necessary when suffering from \textit{extreme} slow data missing. 
As shown in \figref{fig:POIN}, \sysname improves 54.3\% mAP and 55.9\% NDS compared to without data imputation under data missing rates at 75\%. 
Comparatively, without \sysname's data imputation module, a 75\% data missing rate causes an 78.6\% decline in accuracy and a 77.2\% drop in NDS, making the system unusable.

\textbf{Summary.}
Under various asynchrony levels (\ie slow data missing rates), \sysname maintains system effectiveness by preventing accuracy declines. 
Particularly in extreme asychronous cases, \sysname avoids drastic accuracy drop.
% that could render the system inoperative.

% AdaFlow demonstrates robustness to data asynchrony across both BEVFusion and PointPillars models. 
% This capability allows AdaFlow to be effectively applied in scenarios such as autonomous or assisted driving, where point cloud data is often incomplete and the available multi-modal data options are limited to RGB data and point clouds.

\begin{figure}[t]
	\centering
	\includegraphics[width=0.85\linewidth]{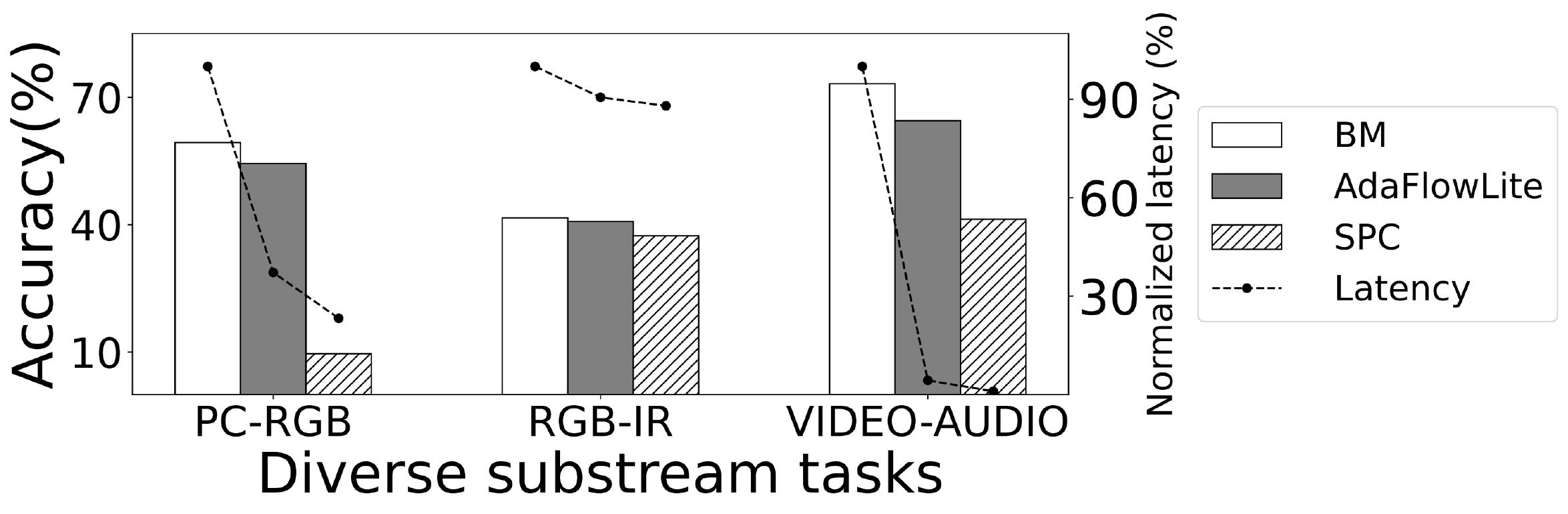}
 \vspace{-3mm}
	\caption{Performance over diverse downstream tasks.}
 \vspace{-3mm}
	\label{fig:gen}
\end{figure}

\subsection{Generalizing to Diverse Input Data}
We evaluate \sysname's generalization ability on three diverse tasks, \ie BEVFusion (\(T_1\)), and two LLM-based tasks ($T_{3}, T_{4}$), across different data missing rates of slow data. 
As shown in \figref{fig:gen}. \textit{First}, \sysname can adapt to various multi-modal tasks.
In detail, on \(T_1\), \(T_3\), and \(T_4\), \sysname achieve 44.8\%, 3.4\%, and 23.2\% accuracy improvements compared to SPC, a typical asynchronous baseline.
\textit{Second}, \sysname performs best when the data volume ratio between fast and slow modalities is large. 
As shown in \figref{fig:acc1}.
For data ratios of 1:4, 8:9, and 1:70, \sysname reduces latency by 62.8\%, 9.4\%, and 95.6\%, respectively, compared to BM, a synchronous method with upper-bound accuracy.
These gains are attributed to the one-fit-all MMLST for dynamic data imputation and leverage a class attention mechanism for maximum data imputation.

% \textit{Second}, \sysname shows improvement over the asynchronous baseline for these downstream tasks, with performance gains of  , under their corresponding data missing rates of slow data. The results demonstrate that \sysname is not a customized system, but rather a general-purpose system.

\subsection{Performance of Affinity Matrix}

\subsubsection{Validity Verification}
\label{val}

\begin{figure}[t]
	\centering
	\includegraphics[width=0.73\linewidth]{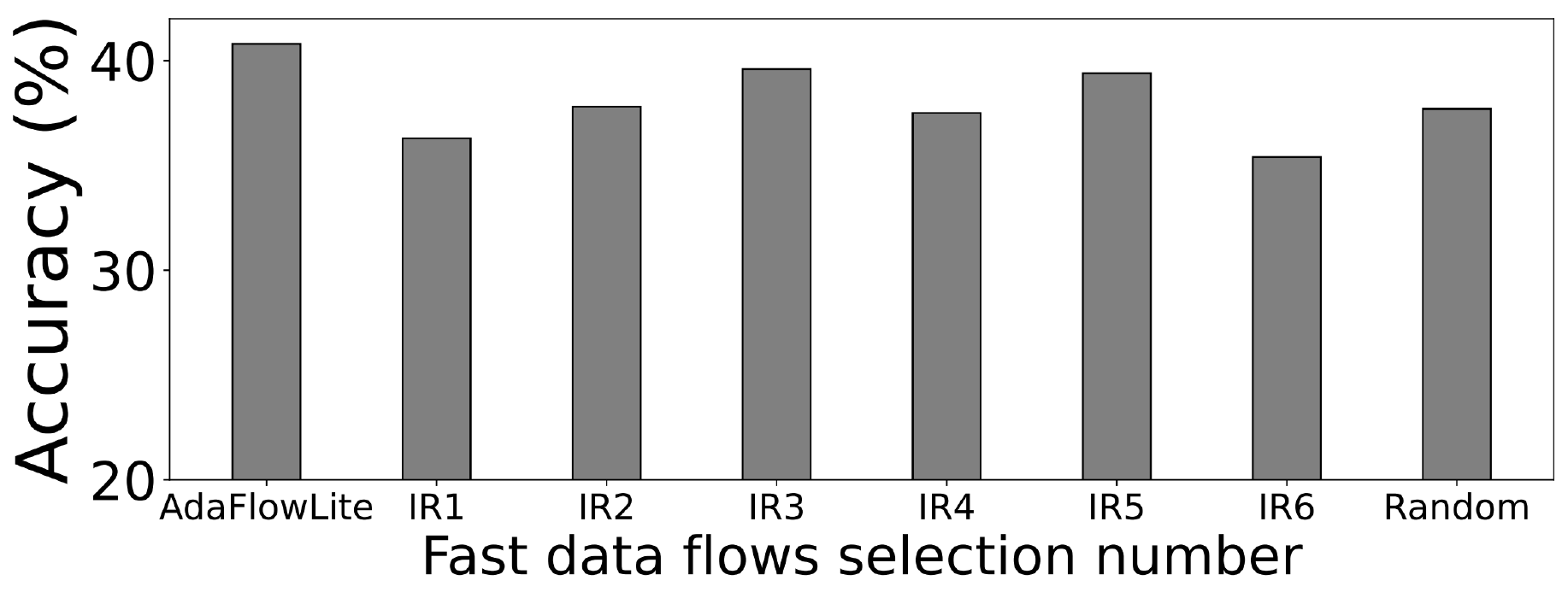}
         \vspace{-3mm}
	\caption{Comparison between \sysname and Swin-T block.}
         \vspace{-3mm}
	\label{fig:acc}
\end{figure}

\begin{figure}[t]
	\centering
	\includegraphics[width=0.73\linewidth]{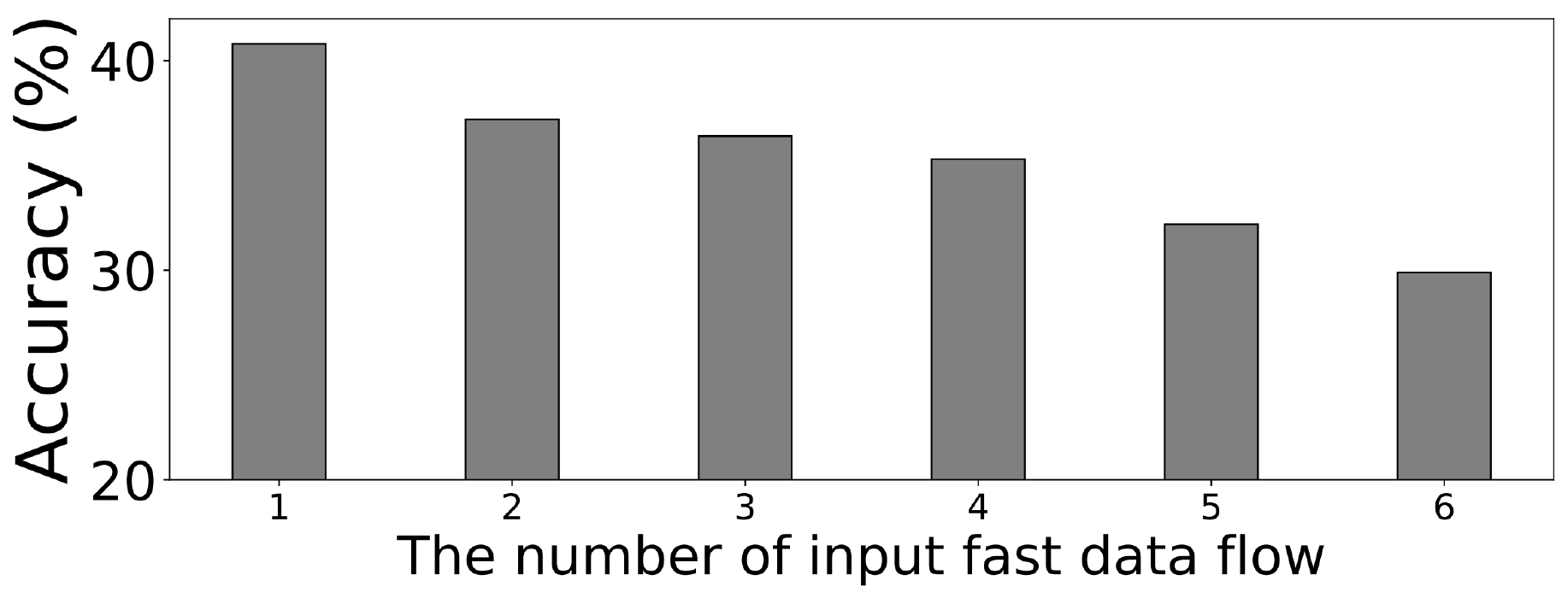}
         \vspace{-3mm}
	\caption{Performance on diverse modality numbers.}
         \vspace{-3mm}
	\label{fig:acc1}
\end{figure}

We test \sysname's affinity matrix on LLM task (\(T_3\)) under 10\% data missing rates of slow data. 
\textit{First}, \sysname achieves higher accuracy than Swin-T block's 6 fixed selections, as shown in \figref{fig:acc}. \sysname's accuracy is 40.8\% and 6 Swin-T block's 6 fixed selections is 36.3\%, 37.8\%, 39.6\%, 37.5\%, 39.4\% and 35.4\%, respectively. The inference derived by \sysname's generalized affinity matrix is also higher than Swin-T block's random selection (37.7\%). 
\textit{Second}, 
\sysname achieves higher accuracy with just one fast modality input, as shown in \figref{fig:acc1}, while accuracy drops to 29.9\% when all modalities are used together.
This proves that the affinity matrix can always select the most suitable combination of available (fast) data for missing (slow) data imputation in diverse situations and supports the argument in \secref{graph}.

\subsubsection{Comparison to Monte Carlo}
\begin{figure}[t]
    \centering 
     \vspace{-2mm}
    \subfloat[1st frame]{\label{fig:t1g}
    \includegraphics[width=0.33\linewidth]{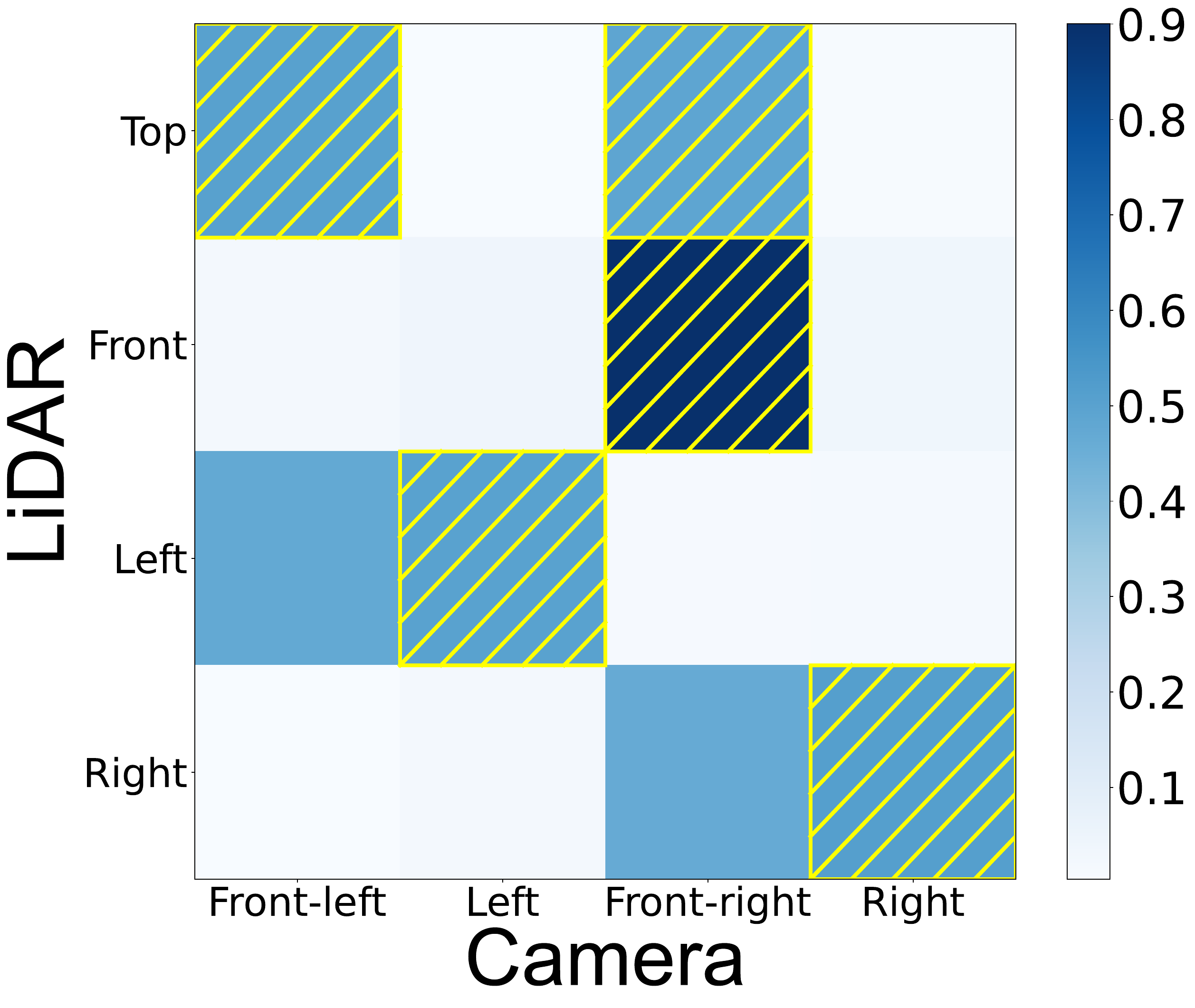}}
     \subfloat[10th frame]{\label{fig:t2g}
    \includegraphics[width=0.33\linewidth]{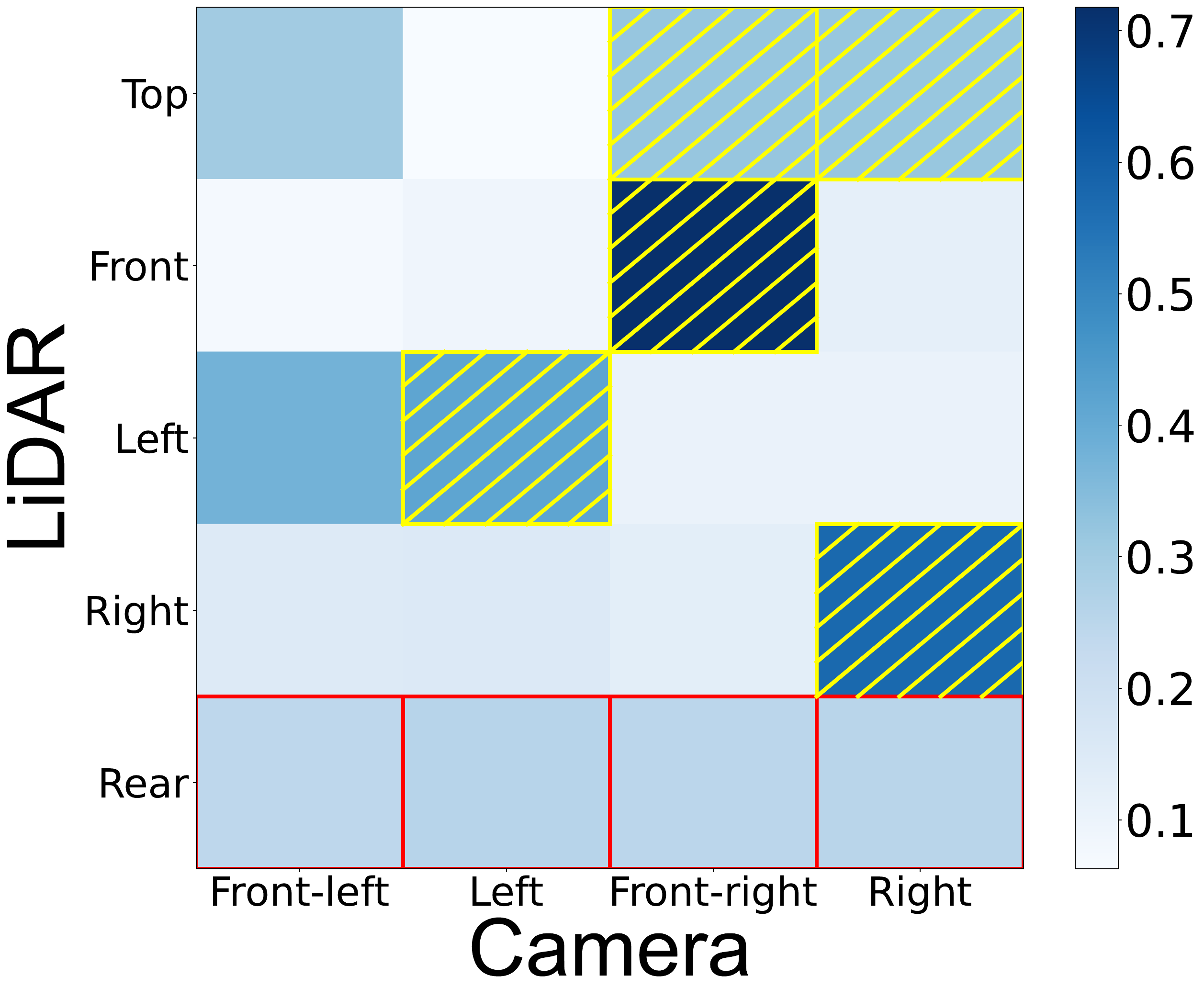}}
    \subfloat[20th frame]{\label{fig:t3g}
    \includegraphics[width=0.33\linewidth]{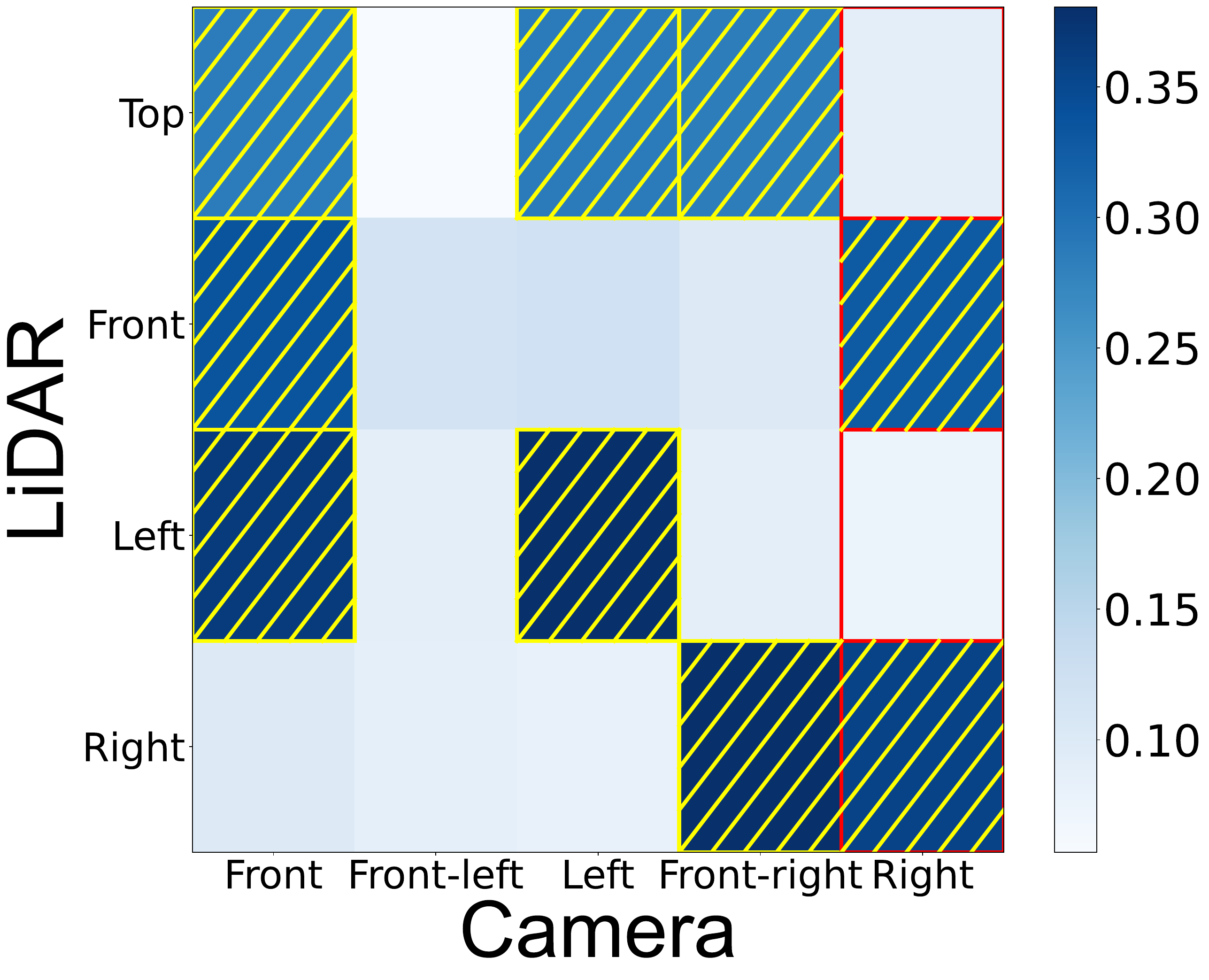}}
     % \vspace{-3mm}
    \caption{\wfm{Visualization of \sysnameposs affinity matrix.}}
     \vspace{-3mm}
\label{fig:vm}
\end{figure}

\begin{figure}[t]
    \centering 
    \vspace{-2mm}
    \subfloat[1st frame]{\label{fig:t1}
    \includegraphics[width=0.33\linewidth]{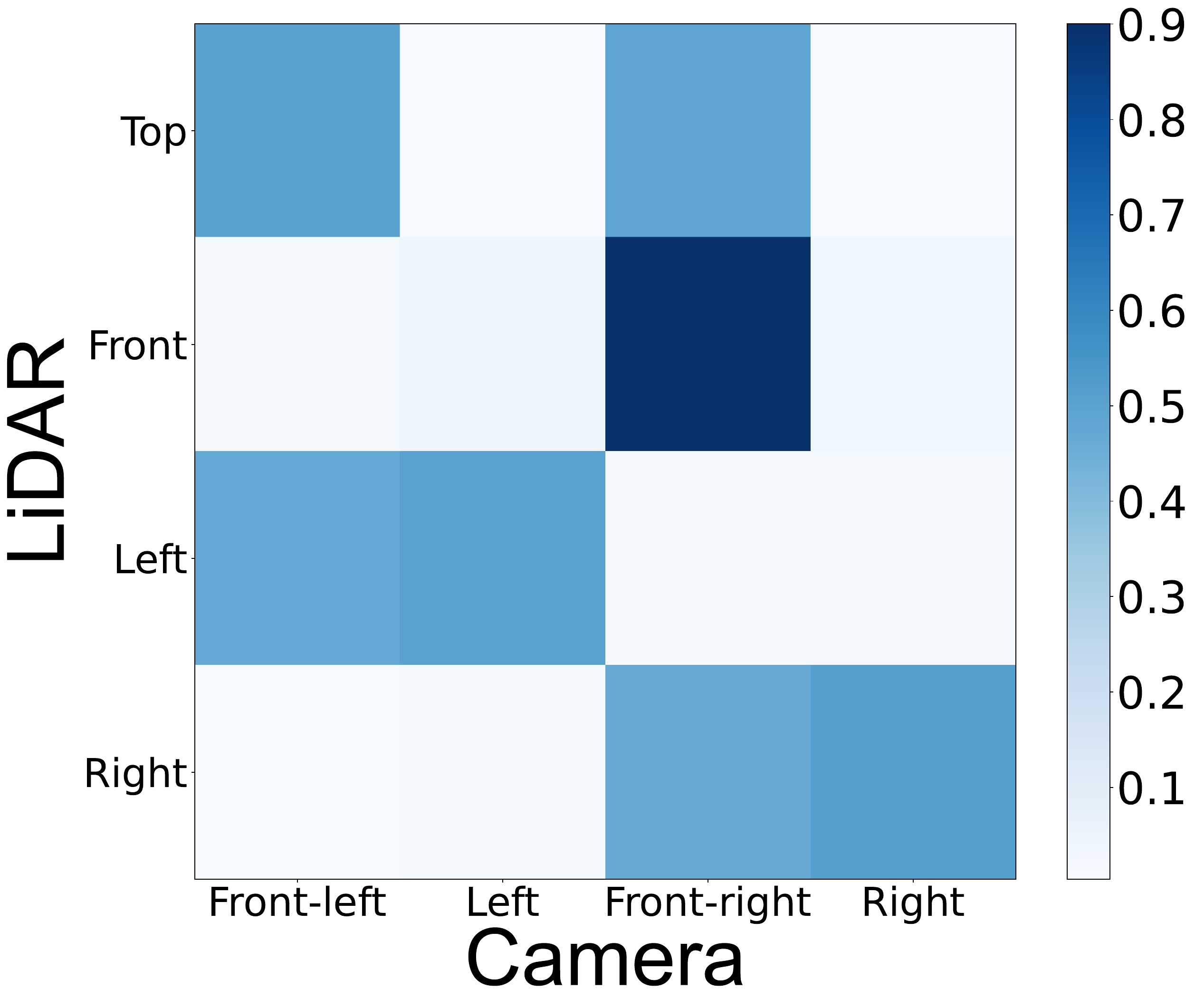}}
     \subfloat[10th frame]{\label{fig:t2}
    \includegraphics[width=0.33\linewidth]{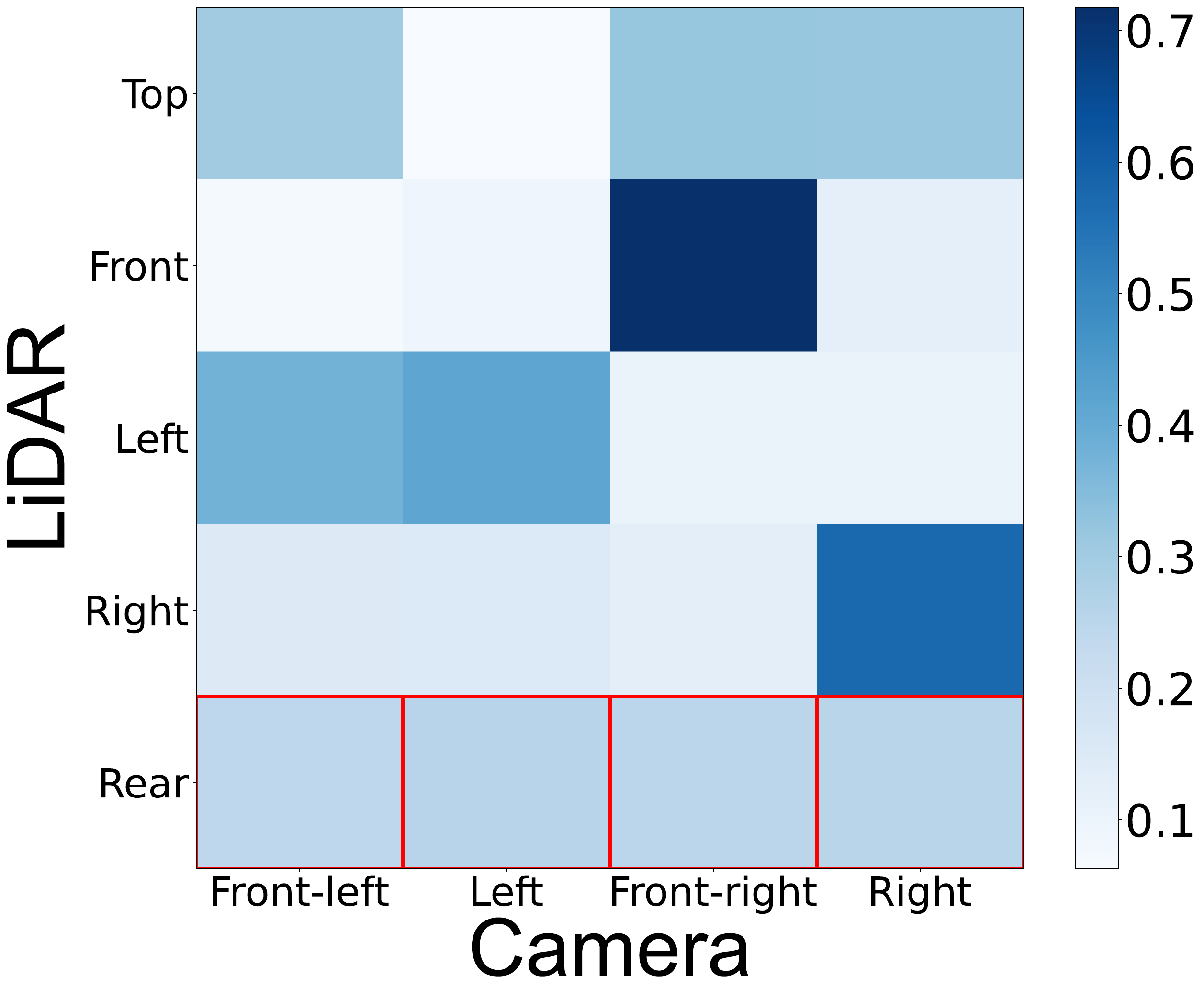}}
    \subfloat[20th frame]{\label{fig:t3}
    \includegraphics[width=0.33\linewidth]{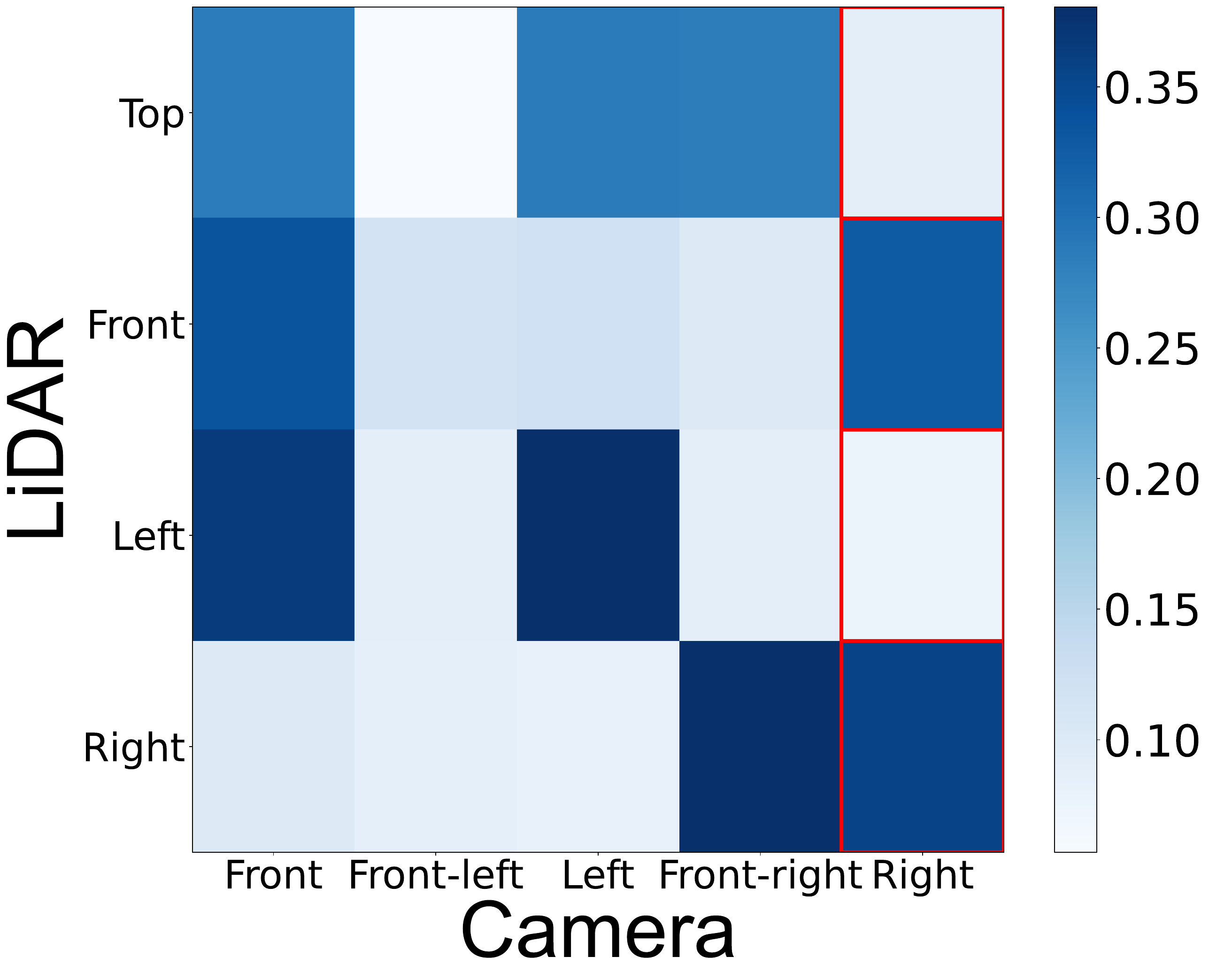}
    \label{fig:gt:c}
    }
     \vspace{-3mm}
    \caption{\wfm{Visualization of expected affinity matrix.}}
     \vspace{-3mm}
\label{fig:gt}
\end{figure}

We compare the affinity matrix outperformed by \sysname and Monte Carlo in \figref{fig:vm} and \figref{fig:gt}. 
For each matrix, the x-axis represents different camera views, the y-axis represents different LiDAR views. 
And elements in the matrix are the affinity value between a LiDAR view and a camera view.
In particular, we use the Monte Carlo~\cite{bib:mc:v} method here to establish relationships between modalities,  which can be regarded as the baseline (\ie ground truth). 
We chose the Monte Carlo method for its efficiency in approximating complex problems through repeated random sampling, achieving reliable solutions quickly.

\wfm{To simulate mobile scenarios where sensors need to be added, we mask the rear Lidar and front camera in the Waymo dataset in the first frame, and add the rear Lidar and front camera in the 10-th frame and the 20-th frame, respectively.
\textit{First}, \sysname's affinity matrix accurately captures the relationships between modalities as shown in \figref{fig:t1} and\figref{fig:t1g}.
And \sysname's affinity matrix can dynamically adapt to changes in inter-modal relationships, as shown in \figref{fig:vm}, the same modal settings exhibited different inter-modal relationships at different times.
These features confirm that \sysnameposs affinity matrix matches the functionality of AdaFlow in handling dynamic modal environments.
\textit{Second}, \sysname's affinity matrix demonstrates exceptional scalability by dynamically accommodating new sensors. 
As shown in \figref{fig:vm} and \figref{fig:gt}, the matrix initially comprises four rows and four columns. 
Integrating the slow sensor rear Lidar by the 10th frame expands the matrix to five rows and four columns. Subsequently, with the addition of the fast sensor front camera by the 20th frame, the matrix adjusts again to four rows and five columns. 
This ability to adapt its structure with introducing new sensors highlights the robustness and flexibility of \sysname in diverse mobile scenarios.}
% \todo{explain why, to show the detail advantages}

% \wfm{We take the dataset waymo as a visualization case. We select three different time points in the first frame, tenth frame and twentieth frame of the dataset to make relation and matrix visualization to concretize its characteristics discussed in the Sec \ref{sec:matrix} and Sec \ref{sec:inference}. As shown in \figref{fig:vm}, for each matrix, the row represents 5 camera views, which are front, front-left, left, front-right, and right, and the column represents 5 LiDAR views, which are top, front, left, right, and rear, the elements represent the affinity between the different modalities. Yellow means selected, the number is obtained from Sec \ref{graph}, that is, top LiDAR selects 3 camera views, non-top LiDAR except rear select 2 camera views and rear LiDAR selects 0 camera view. It can be observed that with the different environment perceived by the sensor in each frame, the affinity matrix changes dynamically, so does the selection. Notably, the rear LiDAR and 5 camera views do not intersect FOV, so there is little difference in the value after standardization, which leads to zero selection.}

\subsection{Case Study}
\figref{fig:flow} illustrates the latency breakdown in conventional end-to-end multi-modal object detection systems for autonomous driving, categorized into data pre-processing (\eg handling or imputing slow data) and executing inference. 
\sysname enhances system performance by utilizing affinity-aware data imputation for pre-processing, thereby reducing the waiting time associated with slow data and improving overall detection responsiveness.
To evaluate \sysname, we generated a 10,000-frame autonomous driving dataset using the Carla\cite{bib:carla:d} autopilot simulator. 
As demonstrated in \figref{fig:case}, the system processes inputs from six camera views (fast data) and one LiDAR view (slow data). 
We benchmark \sysname against three asynchronous methods, \ie AdaFlow, KNN, and PCN, comparing data pre-processing latency and imputation quality. 
We test data quality using Maximum Mean Discrepancy (MMD) and Chamfer Distance (CD) error metrics, comparing the imputed data with actual sensor data.

\figref{fig:c3} shows the results.
\sysname reduces the overall latency of the pre-processing phase from 0.346s, 0.246s, 0.141s, 0.059s to 0.04s, compared to KNN, PCN, BM ,and AdaFlow respectively.
While when imputing data similar to those collected by real sensors (see \figref{fig:c1} and \figref{fig:c2}), \sysname achieves latency reductions of $1.5\sim 8.7\times$ than BM, PCN, KNN, and AdaFlow. 
In practical urban settings, where vehicles commonly travel at 36 km/h, with such latency reduction, \sysname reduces the detection distance interval from 0.59m-3.46m to just 0.4m. 
This significant improvement enhances the timely detection of pedestrians and vehicles, helping accident prevention.

{We deploy \sysname with three different bandwidths, \ie 50bps, 75Mbps, and 100Mbps on NVIDIA AGX Xaxier, similar Mercedes in-car chips. 
As shown in \figref{fig:real}, 
\sysname remains efficient even under low bandwidth. 
Compared to BM, it reduces latency by 49\% at 50Mbps while keeping CD error within acceptable limits (\ie $\leq$ 0.00015).
We also test \sysname's energy cost on the AGX Xavier. As shown in \figref{fig:chips}.
\sysname consumes approximately 15W, which represents a 31\% reduction in energy consumption compared to AdaFlow. This enhanced efficiency makes \sysname more suitable for resource-constrained environments.

\begin{figure}[t]
	\centering
	\includegraphics[width=0.92\linewidth]{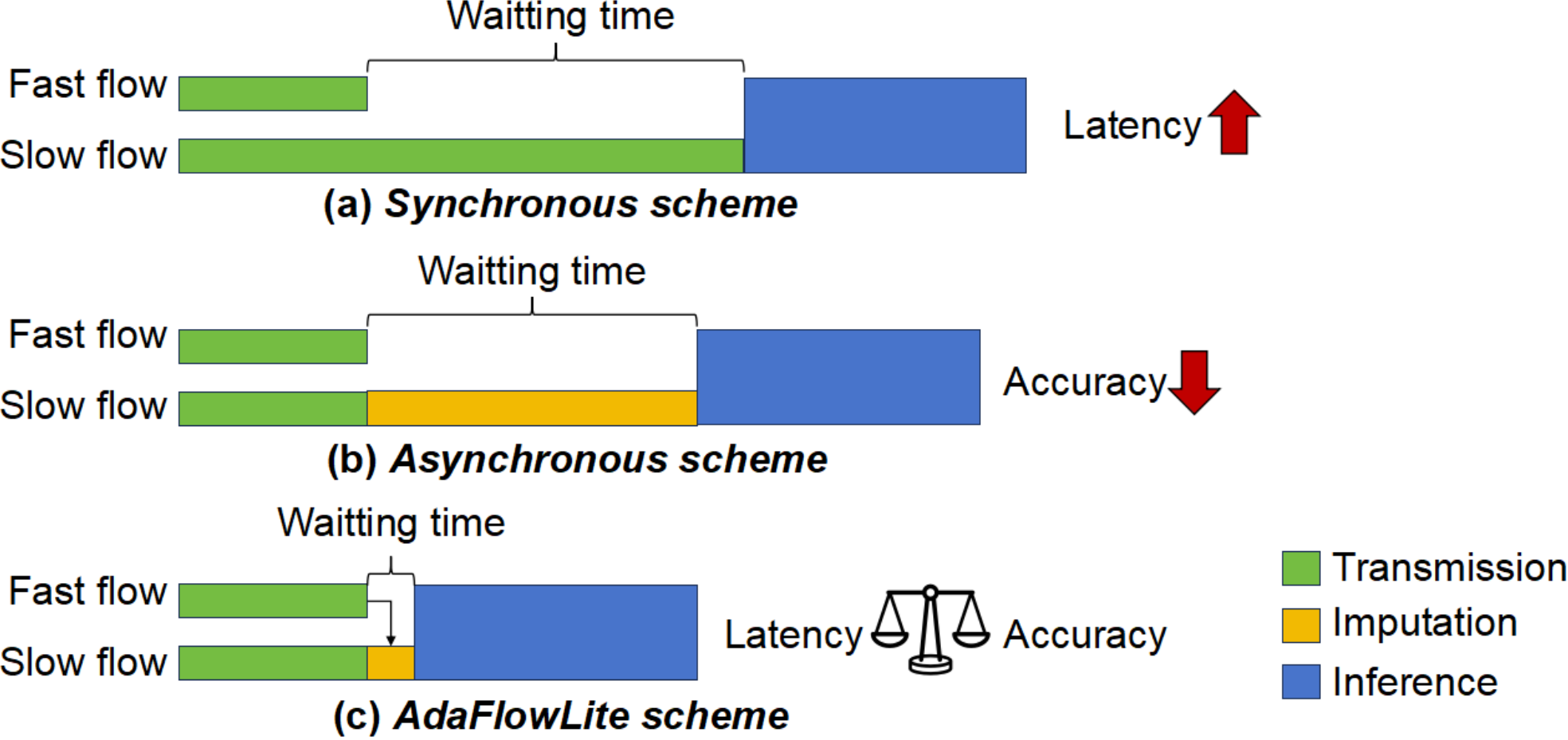}
	\caption{Paradigms of \sysname and baselines.}
	\label{fig:flow}
\end{figure}

\begin{figure}[t]
	\centering
	\includegraphics[width=0.82\linewidth]{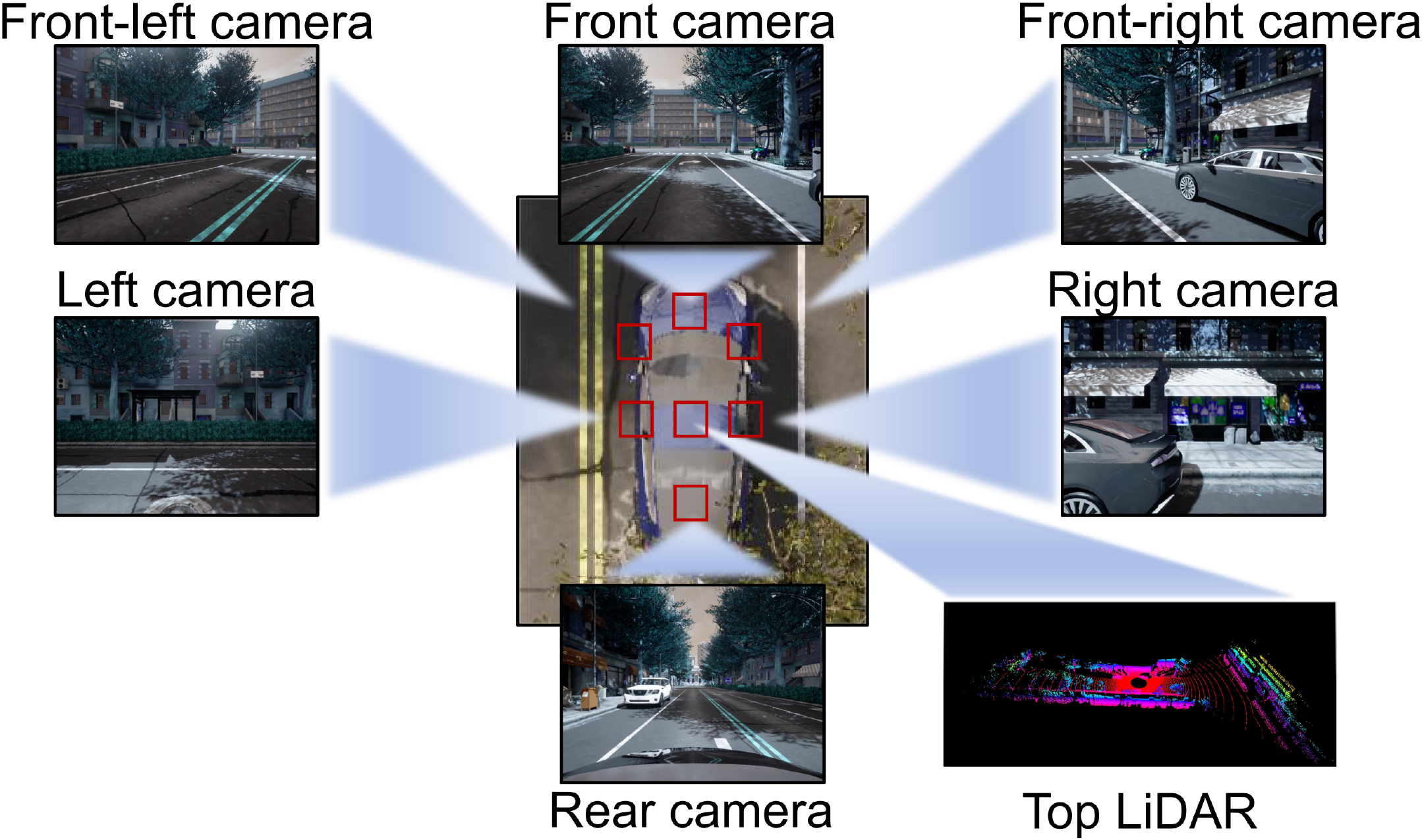}
	\caption{Illustration of seven sensors (six cameras and one LiDAR) on vehicle. }
         % \vspace{-3mm}
	\label{fig:case}
\end{figure}

\begin{figure}[t]
    \centering 
     \vspace{-3mm}
    \subfloat[MMD]{\label{fig:c1}
    \includegraphics[width=0.30\linewidth]{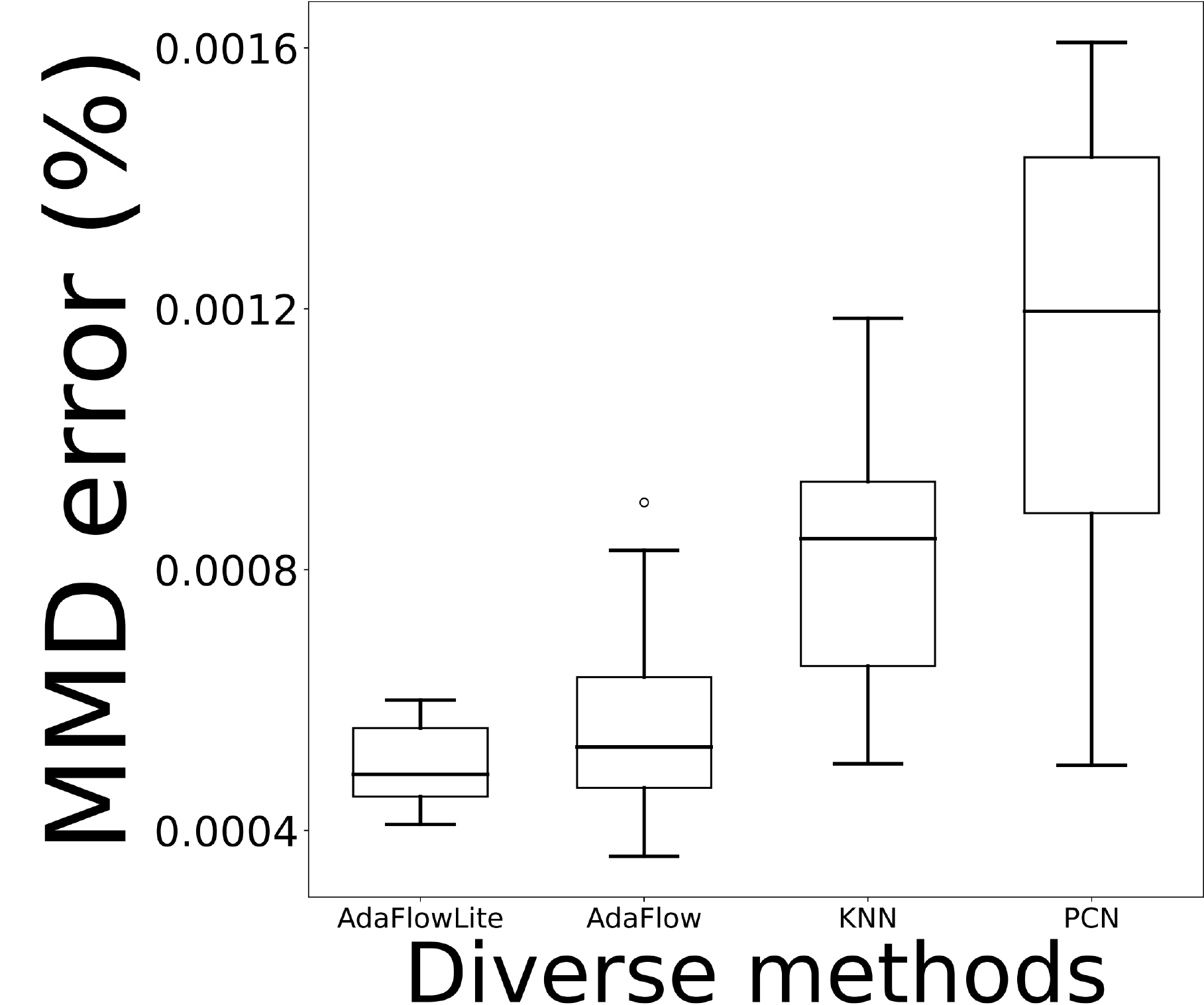}}
     \subfloat[CD]{\label{fig:c2}
     \hspace{2mm}
    \includegraphics[width=0.30\linewidth]{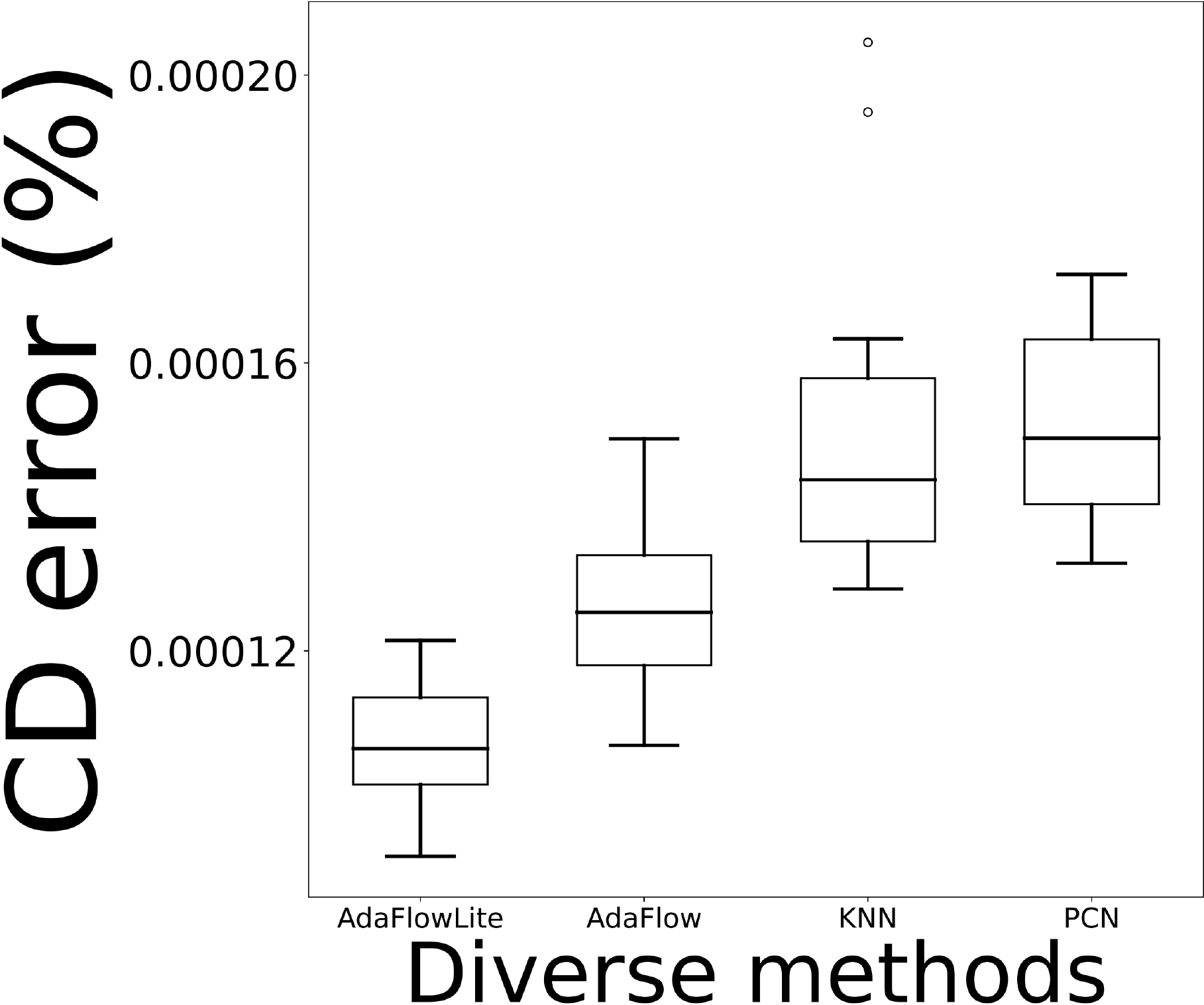}}
    \subfloat[Latency]{\label{fig:c3}
    \hspace{2mm}
    \includegraphics[width=0.30\linewidth]{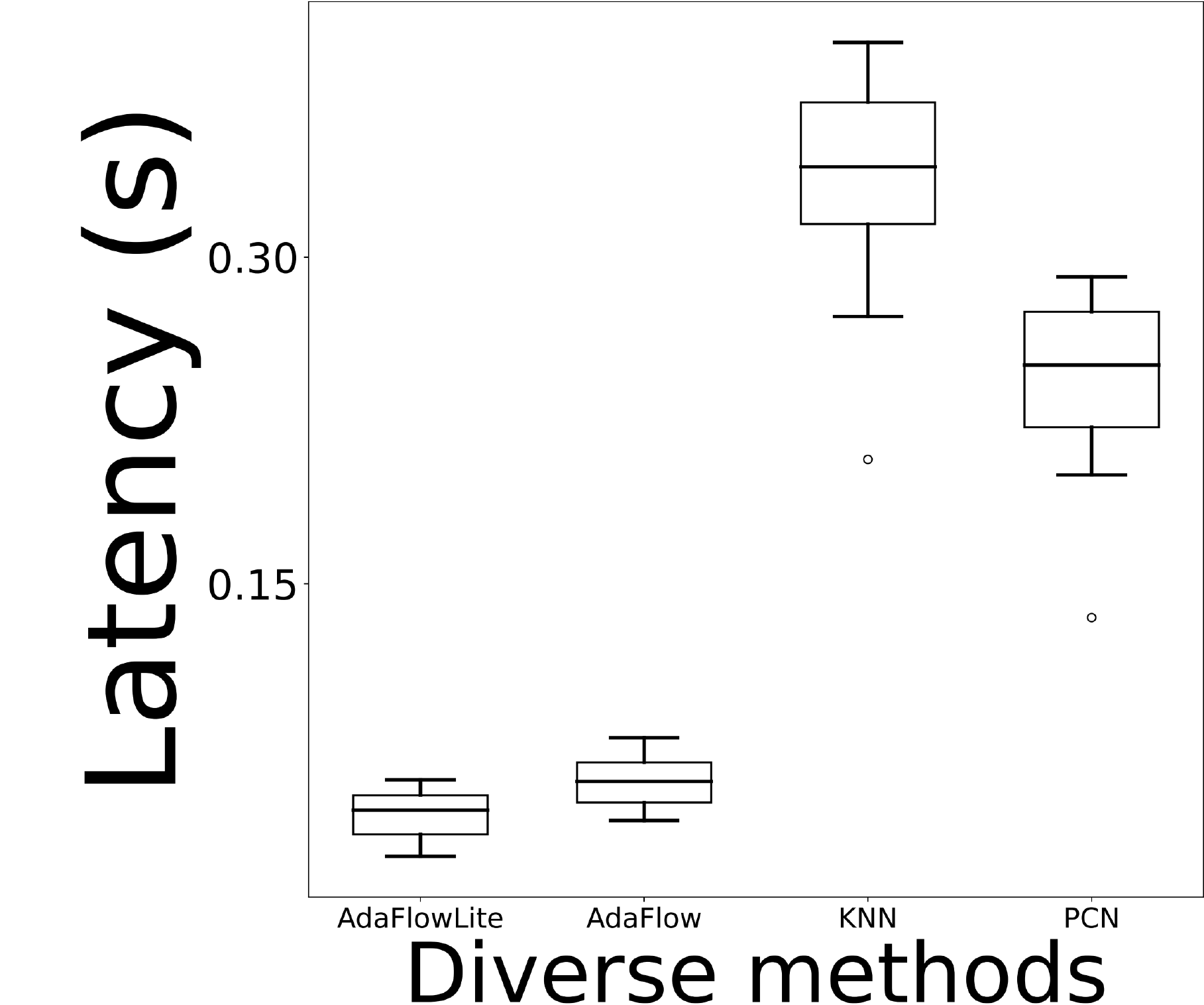}
    }
    % \vspace{-3mm}
    \caption{Comparison between \sysname and asynchronous baselines (AdaFlow, KNN, PCN) in the self-collect dataset under 66\% data missing rates of LiDAR.}
     % \vspace{-3mm}
\label{fig:case1}
\end{figure}

\begin{figure}[t]
	\centering
	\includegraphics[width=0.9\linewidth]{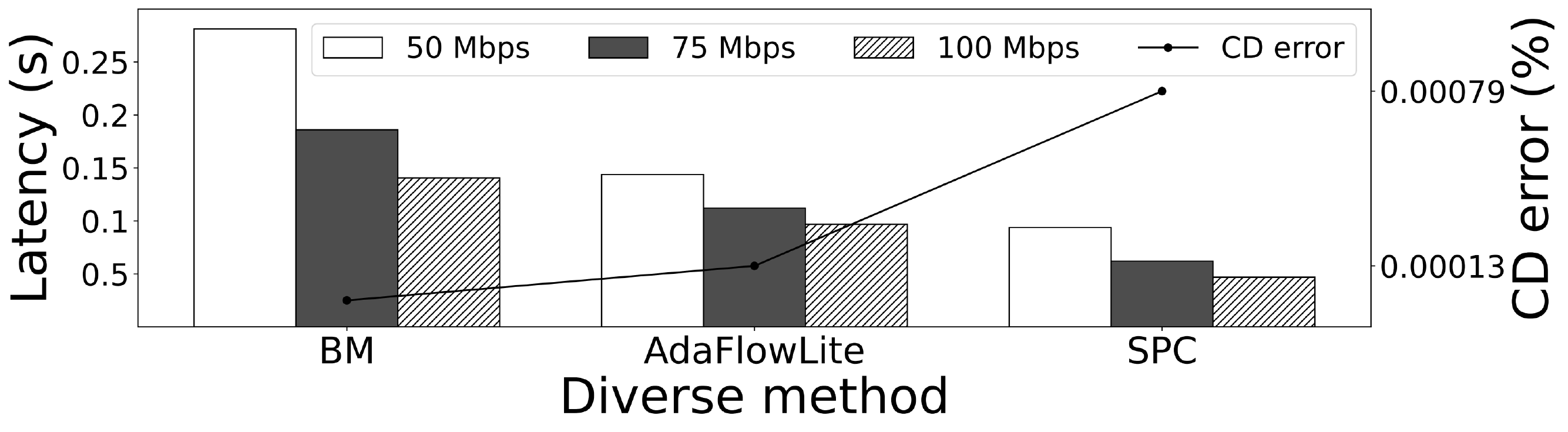}
  \vspace{-3mm}
	\caption{Performance with diverse bandwidths.}
  \vspace{-3mm}
	\label{fig:real}
\end{figure}

 \begin{figure}[t]
 	\centering
 	\includegraphics[width=0.97\linewidth]{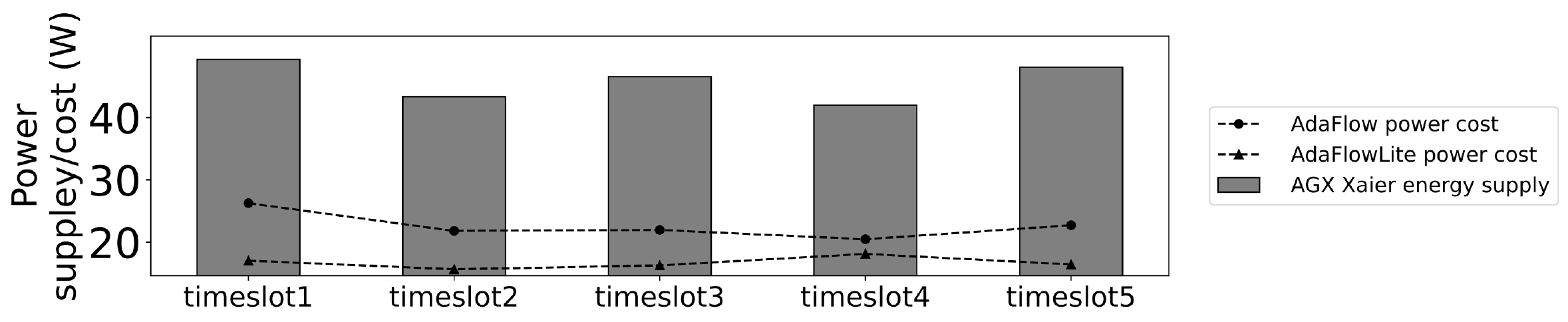}
  \vspace{-3mm}
 	\caption{Energy cost of \sysname on AGX Xavier.}
  \vspace{-3mm}
 	\label{fig:chips}
 \end{figure}

%% file: body/related.tex
\section{Related Work}
\label{sec:related}
\textbf{Multi-modal Inference in Mobile Applications.} 
Multi-modal inference has demonstrated success across a range of mobile sensing tasks such as classification~\cite{bib:multimodal:hong,bib:multimodal:liu,3,bib:echo:l,bib:class:l}, object detection~\cite{4,5,6,bib:spr:l,bib:evo:l,bib:caq:l,bib:surgery:w,bib:adanife:l}, segmentation~\cite{7,8,9}, speech recognition~\cite{10}, autonomous driving~\cite{bib:nuscenses:ca,bib:kitti:g,bib:waymo:s,bib:adashadow:f}, and medical image segmentation~\cite{12}. For instance, the nuScenes dataset~\cite{bib:nuscenses:ca} illustrates how integrating multi-modal data can enhance perception in autonomous driving, improving the effectiveness and user experience in driving and traffic scenarios.
% Multi-modal learning has achieved success on various \lxc{mobile sensing} tasks, including classification~\cite{bib:multimodal:hong,bib:multimodal:liu,3}, object detection~\cite{4,5,6}, segmentation~\cite{7,8,9}, speech recognition~\cite{10}, autonomous driving~\cite{bib:nuscenses:ca,bib:kitti:g,bib:waymo:s}, and medical image segmentation~\cite{12}. 
% For example, The nuScenes~\cite{bib:nuscenses:ca} shows how multi-modal data can enhance the perception of autonomous driving applications, and how multi-modal learning can make mobile applications more powerful, enhancing the user experience in driving and traffic-related scenarios.
% This is the first model to learn features from both the original image and the audio signal and perform joint classification. Experimental results show that the proposed model is significantly better than the audio-based model under high noise conditions.

\textbf{A-/Synchronous Multi-modal Inference.} 
Multi-modal inference processes data from multiple \textit{asynchronous} flows, where the slowest flow often increases waiting time. 
Existing methods are divided into \textit{synchronous blocking} and \textit{asynchronous non-blocking} approaches. 
Synchronous methods, like frame sampling~\cite{bib:fs:z}, wait for all data to arrive, optimizing latency by transmitting keyframes, but risking accuracy if slow data is low quality. 
Asynchronous methods, such as Tianxing \etal~\cite{bib:lls:L}, use predictive completion to reduce wait time but are limited by fixed input modality configurations.
\sysname leverages cross-modal affinity for attention-based fusion, enabling efficient asynchronous inference across diverse input modalities, regardless of number or type.
Based on~\cite{bib:adaflow:wu}, \sysname utilizes MMLST to further optimize the number of parameters so that the framework can run in resource-constrained environment.

% Multi-modal inference processes data from multiple \textit{asynchronously} arriving flows, where the slowest flow often extends the latency. 
% To address this, existing methods are categorized into \textit{synchronous blocking} and \textit{asynchronous non-blocking} approaches. 
% Synchronous blocking methods, such as the frame sampling~\cite{bib:fs:z}, wait for all data to arrive and optimize latency by only transmitting keyframes. 
% However, this can compromise the accuracy if critical data in slower flows are transmitted in low quality. 
% On the other hand, asynchronous non-blocking methods, like the approach by Tianxing \etal~\cite{bib:lls:L}, use predictive completion for slower streams to reduce waiting time and improve latency. 
% Yet, these methods are often coupled with specific numbers and types of input modalities, limiting scalability.
% \lsc{
% \sysname employs cross-modal affinity to realize attention-based fusion and enables efficient asynchronous inference across diverse input modalities, regardless of their number or type.
% }
% \sysname enhances efficiency through an asynchronous non-blocking approach while maintaining accuracy. 
% Also, it utilizes a cross-modal affinity metric as an attention-based fusion mechanism which computes an attention matrix through \textit{affinity calculation} and utilizes this matrix to perform a weighted fusion of features. Such a strategy enables efficient inference across various input modalities, irrespective of their number and type.

\textbf{Cross-modal Affinity Metrics in Multi-modal Inference.} 
The affinity metric identifies which modalities can be omitted without significantly impacting cross-modal inference accuracy. Existing approaches to quantify affinity fall into two categories: \textit{model-based} and \textit{function-based}. 
Model-based methods, like ReID~\cite{bib:reid:lu}, integrate the affinity matrix into the loss function during training, capturing affinity between specific modalities but requiring retraining for each new input. Function-based methods, such as Taskonomy~\cite{bib:taskonomy:AZ}, compute affinity from statistical data features without modifying the model. 
\wfm{\sysname leverages the flexible function-based approach of the preliminary AdaFlow~\cite{bib:adaflow:wu}, and extends its capabilities with a prior-free affinity refinement module for seamlessly integrating new sensors into the structured modality affinity matrix by utilizing \textit{prior-free analysis} specifically designed to effectively manage \textit{unseen} modalities. 
In contrast to existing methods, which are restricted to predefined modalities and lack adaptability in accommodating new sensor types or quantities, \sysname introduces a modality-agnostic and scalable affinity quantification method.
This approach significantly enhances cross-modal inference, particularly in dynamic and asynchronous settings, addressing critical challenges in the fusion and analysis of sensor data.}

%% file: body/conclusion.tex
\section{Conclusion}
To tackle the challenges of data \textit{asynchrony} and \textit{heterogeneity} in distributed mobile environments, this paper presents \sysname, a system designed to perform inference as soon as partial asynchronous data becomes available, thereby enhancing operational efficiency.
\wfm{\textit{First}, \sysname introduces a generalized modality-scalable affinity matrix to model and optimize inference processes, enabling dynamic adaptation across mobile scenarios of inputs and facilitating effective multi-modal fusion. Additionally, \sysname employs a prior-free affinity refinement module for seamlessly integrating \textit{unseen} modalities.}
\textit{Second}, \sysname employs a one-fit-all multi-modal lightweight Swin Transformer (MMLST) that is specifically designed for high-accuracy and low-latency inference. This model is not only lightweight, enabling minimal computational overhead, but also non-blocking, which facilitates continuous processing across various input modalities without the need for retraining. This lightweight architecture is crucial for maintaining performance efficiency in resource-constrained mobile environments.
Through evaluations on real-world multi-modal tasks, \sysname has demonstrated significant reductions in inference latency and substantial improvements in accuracy, all while maintaining a low resource footprint. Looking forward, there is potential to expand \sysname's capabilities to include adaptive affinity-based data imputation across a swarm of IoT devices.

% Future improvements can focus on real-time multi-modal inference on mobile devices, minimizing computational overhead through adaptive, affinity-based strategies while maintaining accuracy.

% To address the challenges of data asynchrony and heterogeneity in distributed mobile environments, this paper introduces \sysname. 
% It performs inference as soon as partial asynchronous data is available, thus enhancing data processing efficiency.
% \textit{First}, \sysname 
% employs a computational strategy that models and optimizes inference based on the generalized modality affinity matrix and dynamically adapts to dynamic and diverse inputs for effective multi-modal fusion. 
% Furthermore, \sysname uses the analytic hierarchy process (AHP) to normalize matrix values.
% \textit{Second}, \sysname adopts a one-fit-all affinity attention-based conditional GAN (ACGAN) for high-accuracy, low-latency, non-blocking inference that adapts to various input modalities without retraining. 
% We test \sysname across various multi-modal tasks characterized by real-world data heterogeneity and asynchrony, achieving substantial reductions in inference time and enhancements in accuracy.
% Future enhancements to \sysname will enable real-time multi-modal inference on mobile devices. 
% We can reduce computational overhead through adaptive, affinity-based non-blocking strategies, ensuring efficient performance on mobile platforms without sacrificing accuracy.